\documentclass{article}

\usepackage{iclr2027_conference,times}

\usepackage{amsmath,amsfonts,bm}

\def\eqref#1{equation~\ref{#1}}

\def\1{\bm{1}}

\DeclareMathAlphabet{\mathsfit}{\encodingdefault}{\sfdefault}{m}{sl}
\SetMathAlphabet{\mathsfit}{bold}{\encodingdefault}{\sfdefault}{bx}{n}

\usepackage{booktabs}
\usepackage{longtable}
\usepackage{graphicx}
\usepackage{caption}
\usepackage{float}
\usepackage{placeins}
\usepackage{hyperref}
\usepackage{microtype}
\usepackage{url}
\usepackage[table]{xcolor}
\newcolumntype{P}[1]{>{\raggedright\arraybackslash}p{#1}}
\definecolor{EHRInk}{HTML}{233348}
\definecolor{EHRRule}{HTML}{DCE3EC}
\definecolor{EHRGen}{HTML}{426AA0}
\definecolor{EHRExt}{HTML}{258C83}
\definecolor{EHRCls}{HTML}{8570AD}
\definecolor{EHRRet}{HTML}{BC8938}
\definecolor{EHRSTS}{HTML}{BE6E7B}
\definecolor{EHRRank}{HTML}{638850}
\newcommand{\caplabel}[2]{\begingroup\setlength{\fboxsep}{2.5pt}\colorbox{#1!12}{\textcolor{#1}{\makebox[23pt][c]{\bfseries #2}}}\endgroup}
\newcommand{\caphead}[2]{\cellcolor{#1!12}\textcolor{#1}{\bfseries #2}}

\usepackage{xspace}

\newcommand{\ehrbench}{\textsc{EHR-Bench}\xspace}

\newcommand{\OpenRecordCount}{8,336}

\newcommand{\TableOpenCategoryRows}{%
Claude-opus-5 & \textbf{74.53} & 55.03 & \textbf{92.02} & 71.20 & \textbf{63.37} & 80.10 & \textbf{85.45} \\
Kimi-K3 & \underline{74.28} & \underline{55.16} & 91.66 & 71.23 & \underline{61.84} & 81.38 & 84.42 \\
GPT-5.6-sol & 74.10 & 54.56 & \underline{91.68} & \textbf{72.14} & 60.84 & 80.05 & 85.32 \\
Gemini-3.5-flash & 74.01 & 53.00 & 90.31 & \underline{72.10} & 60.54 & \textbf{82.69} & \underline{85.41} \\
Claude-opus-4.8 & 73.81 & 54.92 & 91.36 & 71.51 & 60.44 & 79.84 & 84.81 \\
GLM-5.3 & 73.54 & \textbf{57.30} & 91.57 & 71.41 & 59.95 & 81.92 & 79.11 \\
GPT-5.5 & 73.28 & 54.24 & 89.19 & 71.89 & 60.96 & 79.15 & 84.25 \\
Qwen3.8-max & 72.88 & 54.14 & 90.79 & 71.34 & 61.66 & 75.98 & 83.39 \\
GLM-5.2 & 72.50 & 52.35 & 89.49 & 71.56 & 61.08 & \underline{82.13} & 78.36 \\
Kimi-2.6 & 72.15 & 50.76 & 87.93 & 69.95 & 60.47 & 81.59 & 82.18 \\
DeepSeek v4-Pro & 70.94 & 49.84 & 88.86 & 65.86 & 59.47 & 80.86 & 80.72 \\
MiniMax-M3 & 69.03 & 40.06 & 86.99 & 67.00 & 60.37 & 78.82 & 80.96 \\%
}

\newcommand{\TableOpenTaskRows}{%
Claude-opus-5 & 45.81 & 64.25 & \textbf{92.02} & 79.35 & 65.53 & 68.71 & \textbf{63.37} & 60.80 & 81.11 & \textbf{98.39} & \underline{73.72} & \textbf{97.19} \\
Kimi-K3 & 44.63 & 65.68 & 91.66 & 78.48 & 65.53 & 69.68 & \underline{61.84} & 68.04 & 80.70 & 95.40 & 72.78 & 96.06 \\
GPT-5.6-sol & 43.84 & 65.28 & \underline{91.68} & 80.26 & \underline{66.15} & 70.01 & 60.84 & 61.23 & \underline{81.44} & \underline{97.47} & 73.49 & \underline{97.16} \\
Gemini-3.5-flash & 44.62 & 61.39 & 90.31 & 79.87 & 65.53 & \underline{70.91} & 60.54 & \underline{70.71} & 80.93 & 96.44 & \textbf{74.26} & 96.56 \\
Claude-opus-4.8 & 44.03 & \textbf{65.82} & 91.36 & 79.51 & \textbf{66.46} & 68.56 & 60.44 & 63.86 & 81.32 & 94.35 & 73.66 & 95.96 \\
GLM-5.3 & \textbf{48.86} & \underline{65.75} & 91.57 & 79.64 & 64.60 & 69.98 & 59.95 & 68.74 & 80.81 & 96.21 & 73.30 & 84.92 \\
GPT-5.5 & 43.21 & 65.27 & 89.19 & \textbf{81.09} & 64.91 & 69.66 & 60.96 & 60.22 & 81.02 & 96.20 & 71.36 & 97.13 \\
Qwen3.8-max & \underline{46.13} & 62.14 & 90.79 & \underline{81.08} & 64.91 & 68.02 & 61.66 & 56.39 & 79.86 & 91.68 & 72.60 & 94.18 \\
GLM-5.2 & 45.84 & 58.86 & 89.49 & 77.93 & 65.84 & \textbf{70.92} & 61.08 & 69.17 & \textbf{81.47} & 95.75 & 69.74 & 86.98 \\
Kimi-2.6 & 44.52 & 56.99 & 87.93 & 75.39 & 64.91 & 69.54 & 60.47 & 70.07 & 80.70 & 94.01 & 70.07 & 94.29 \\
DeepSeek v4-Pro & 45.61 & 54.08 & 88.86 & 72.02 & 59.94 & 65.62 & 59.47 & \textbf{73.42} & 79.68 & 89.48 & 67.32 & 94.13 \\
MiniMax-M3 & 43.85 & 36.26 & 86.99 & 76.36 & 58.70 & 65.95 & 60.37 & 68.24 & 79.47 & 88.74 & 68.86 & 93.07 \\%
}

\newcommand{\TablePrivateCategoryRows}{%
Kimi-K3 & \textbf{70.48} & 51.34 & 92.51 & 70.05 & 76.32 & \textbf{67.56} & \textbf{65.12} \\
Qwen3.8-max & 70.18 & 51.55 & \textbf{92.95} & \textbf{70.51} & 77.36 & 64.49 & 64.21 \\
GLM-5.2 & 70.14 & \textbf{52.51} & 92.71 & 68.53 & \textbf{77.79} & 66.19 & 63.13 \\
Kimi-2.6 & 69.68 & 51.00 & 90.54 & 69.83 & 75.77 & 65.97 & 64.99 \\
DeepSeek v4-Pro & 68.55 & 47.75 & 91.94 & 67.51 & 72.88 & 67.34 & 63.89 \\
Internal solution & 63.79 & 47.53 & 84.24 & 69.28 & 69.86 & 48.14 & 63.70 \\%
}

\newcommand{\TablePrivateTaskRows}{%
Kimi-K3 & 48.96 & 53.73 & 92.51 & 72.83 & 72.98 & 64.34 & 76.32 & 49.82 & 63.55 & 89.31 & 67.05 & 63.20 \\
Qwen3.8-max & 49.80 & 53.30 & 92.95 & 72.98 & 72.80 & 65.76 & 77.36 & 39.82 & 64.96 & 88.68 & 65.88 & 62.54 \\
GLM-5.2 & 52.09 & 52.93 & 92.71 & 73.36 & 69.61 & 62.63 & 77.79 & 43.32 & 65.86 & 89.40 & 65.21 & 61.06 \\
Kimi-2.6 & 51.47 & 50.52 & 90.54 & 74.27 & 70.36 & 64.87 & 75.77 & 59.06 & 57.63 & 81.23 & 67.53 & 62.45 \\
DeepSeek v4-Pro & 50.99 & 44.50 & 91.94 & 72.79 & 68.48 & 61.26 & 72.88 & 51.91 & 67.25 & 82.85 & 64.43 & 63.34 \\
Internal solution & 45.70 & 49.37 & 84.24 & 63.42 & 70.12 & 74.30 & 69.86 & 47.27 & 40.81 & 56.34 & 61.43 & 65.98 \\%
}

\newcommand{\TableCalibrationRows}{%
Overall & 1.00 & 10/10 & 0 & 0 \\
Generation & 0.60 & 7/10 & 3 & 0 \\
Extraction & 0.70 & 8/10 & 2 & 0 \\
Classification & 0.40 & 7/10 & 3 & 0 \\
Retrieval & 0.60 & 7/10 & 3 & 0 \\
STS & 0.20 & 6/10 & 4 & 0 \\
Ranking & 0.90 & 9/10 & 1 & 0 \\%
}

\newcommand{\TableTaskComparisonRows}{%
JDG & 0.00 & 5 & 5 & 0 \\
PJFA & 1.00 & 10 & 0 & 0 \\
QP & 0.70 & 8 & 2 & 0 \\
JC & 0.10 & 5 & 5 & 0 \\
TC & 0.36 & 6 & 3 & 1 \\
PJF & 0.00 & 4 & 6 & 0 \\
TR & 0.60 & 7 & 3 & 0 \\
SimT & 0.80 & 8 & 2 & 0 \\
SimJ & -0.36 & 3 & 6 & 1 \\
Dedup & 0.70 & 8 & 2 & 0 \\
J2T & 0.70 & 8 & 2 & 0 \\
T2J & 0.30 & 6 & 4 & 0 \\%
}

\newcommand{\TableSourceRows}{%
public\_dataset & 695 & 8.3 \\
public\_job\_descriptions & 1,505 & 18.1 \\
public\_data\_derived & 3,732 & 44.8 \\
controlled\_synthetic & 1,690 & 20.3 \\
synthetic & 714 & 8.6 \\%
}

\newcommand{\TableGenerationComponentRows}{%
Claude-opus-5 & 17.01 & 89.01 & 45.81 & 80.96 & 17.24 & 92.96 & 64.25 \\
Kimi-K3 & 20.01 & 81.54 & 44.63 & 83.60 & 20.03 & 89.34 & 65.68 \\
GPT-5.6-sol & 19.70 & 80.04 & 43.84 & 80.05 & 22.25 & 92.93 & 65.28 \\
Gemini-3.5-flash & 16.54 & 86.74 & 44.62 & 75.46 & 19.26 & 89.43 & 61.39 \\
Claude-opus-4.8 & 20.43 & 79.43 & 44.03 & 84.40 & 20.67 & 87.09 & 65.82 \\
GLM-5.3 & 20.69 & 91.12 & 48.86 & 79.82 & 23.05 & 94.64 & 65.75 \\
GPT-5.5 & 19.17 & 79.29 & 43.21 & 80.73 & 21.05 & 92.94 & 65.27 \\
Qwen3.8-max & 19.71 & 85.75 & 46.13 & 76.83 & 18.47 & 90.91 & 62.14 \\
GLM-5.2 & 19.97 & 84.65 & 45.84 & 69.72 & 21.56 & 87.63 & 58.86 \\
Kimi-2.6 & 19.48 & 82.09 & 44.52 & 67.66 & 17.25 & 89.93 & 56.99 \\
DeepSeek v4-Pro & 19.64 & 84.57 & 45.61 & 59.86 & 21.74 & 88.13 & 54.08 \\
MiniMax-M3 & 19.79 & 79.93 & 43.85 & 27.18 & 16.91 & 88.01 & 36.26 \\%
}

\newcommand{\TableHierarchyRows}{%
Claude-opus-5 & 87.4 (1064) & 71.5 (1064) & 87.5 (32) & 71.4 (161) & 59.6 (161) \\
Kimi-K3 & 87.6 (1064) & 69.8 (1064) & 75.0 (32) & 73.3 (161) & 57.8 (161) \\
GPT-5.6-sol & 88.8 (1064) & 71.9 (1064) & 90.6 (32) & 72.7 (161) & 59.6 (161) \\
Gemini-3.5-flash & 87.8 (1064) & 72.1 (1064) & 93.8 (32) & 71.4 (161) & 59.6 (161) \\
Claude-opus-4.8 & 87.7 (1064) & 71.7 (1064) & 78.1 (32) & 72.0 (161) & 60.9 (161) \\
GLM-5.3 & 87.8 (1064) & 71.7 (1064) & 78.1 (32) & 70.8 (161) & 58.4 (161) \\
GPT-5.5 & 88.7 (1064) & 73.6 (1064) & 93.8 (32) & 72.0 (161) & 57.8 (161) \\
Qwen3.8-max & 89.0 (1064) & 73.4 (1064) & 87.5 (32) & 71.4 (161) & 58.4 (161) \\
GLM-5.2 & 86.3 (1064) & 69.8 (1064) & 81.2 (32) & 73.3 (161) & 58.4 (161) \\
Kimi-2.6 & 85.5 (1064) & 65.7 (1064) & 71.9 (32) & 72.7 (161) & 57.1 (161) \\
DeepSeek v4-Pro & 81.7 (1064) & 62.5 (1064) & 84.4 (32) & 68.3 (161) & 51.6 (161) \\
MiniMax-M3 & 86.2 (1063) & 66.9 (1063) & 75.0 (32) & 67.1 (161) & 50.3 (161) \\%
}

\newcommand{\TableRankingValidityRows}{%
Claude-opus-5 & 116/116 & 361/361 \\
Kimi-K3 & 116/116 & 360/361 \\
GPT-5.6-sol & 116/116 & 361/361 \\
Gemini-3.5-flash & 116/116 & 361/361 \\
Claude-opus-4.8 & 116/116 & 360/361 \\
GLM-5.3 & 115/116 & 318/361 \\
GPT-5.5 & 116/116 & 361/361 \\
Qwen3.8-max & 116/116 & 354/361 \\
GLM-5.2 & 113/116 & 326/361 \\
Kimi-2.6 & 116/116 & 359/361 \\
DeepSeek v4-Pro & 116/116 & 361/361 \\
MiniMax-M3 & 116/116 & 356/361 \\%
}

\newcommand{\TablePairedRows}{%
GLM-5.3 & JDG & 941 & 941 & +3.05 & [+2.85, +3.25] \\
GLM-5.3 & PJFA & 872 & 872 & +1.50 & [+0.42, +2.59] \\
GLM-5.3 & QP & 714 & 714 & -0.45 & [-1.66, +0.75] \\
GLM-5.3 & JC & 1064 & 1064 & +0.28 & [-1.41, +2.02] \\
GLM-5.3 & TC & 161 & 161 & -0.93 & [-6.21, +4.04] \\
GLM-5.3 & PJF & 1386 & 874 & +1.27 & [-0.32, +2.85] \\
GLM-5.3 & TR & 184 & 18 & -3.41 & [-4.46, +0.24] \\
GLM-5.3 & SimT & 811 & 274 & +7.94 & [-0.86, +12.03] \\
GLM-5.3 & SimJ & 856 & 618 & -0.30 & [-1.72, +1.16] \\
GLM-5.3 & Dedup & 870 & 762 & -2.18 & [-3.47, -0.95] \\
GLM-5.3 & J2T & 116 & 24 & -0.42 & [-2.90, +2.59] \\
GLM-5.3 & T2J & 361 & 1 & -12.27 & -- \\
DeepSeek & JDG & 941 & 941 & -0.20 & [-0.39, -0.01] \\
DeepSeek & PJFA & 872 & 872 & -10.17 & [-12.04, -8.30] \\
DeepSeek & QP & 714 & 714 & -3.16 & [-4.59, -1.79] \\
DeepSeek & JC & 1064 & 1064 & -7.33 & [-9.45, -5.33] \\
DeepSeek & TC & 161 & 161 & -5.59 & [-10.87, -0.62] \\
DeepSeek & PJF & 1386 & 874 & -3.09 & [-5.20, -0.92] \\
DeepSeek & TR & 184 & 18 & -3.90 & [-5.18, +0.44] \\
DeepSeek & SimT & 811 & 274 & +12.62 & [+1.83, +17.18] \\
DeepSeek & SimJ & 856 & 618 & -1.43 & [-3.36, +0.40] \\
DeepSeek & Dedup & 870 & 762 & -8.91 & [-11.01, -6.88] \\
DeepSeek & J2T & 116 & 24 & -6.40 & [-9.88, -3.01] \\
DeepSeek & T2J & 361 & 1 & -3.06 & -- \\
MiniMax & JDG & 941 & 941 & -1.96 & [-2.17, -1.75] \\
MiniMax & PJFA & 872 & 872 & -27.98 & [-29.76, -26.24] \\
MiniMax & QP & 714 & 714 & -5.03 & [-6.52, -3.55] \\
MiniMax & JC & 1063 & 1063 & -3.07 & [-5.06, -1.05] \\
MiniMax & TC & 161 & 161 & -6.83 & [-12.42, -1.24] \\
MiniMax & PJF & 1386 & 874 & -2.76 & [-5.21, -0.30] \\
MiniMax & TR & 184 & 18 & -3.00 & [-4.09, +0.80] \\
MiniMax & SimT & 811 & 274 & +7.44 & [-3.52, +11.70] \\
MiniMax & SimJ & 855 & 617 & -1.60 & [-3.11, -0.13] \\
MiniMax & Dedup & 870 & 762 & -9.65 & [-11.87, -7.47] \\
MiniMax & J2T & 116 & 24 & -4.86 & [-8.96, -0.80] \\
MiniMax & T2J & 361 & 1 & -4.12 & -- \\%
}

\newcommand{\ReviewSupportBaselineSourceRows}{%
TR & Public & 104 & 16.68 & 19.65 & 37.90 & 45.55 & +7.66 \\
TR & Derived & 80 & 53.34 & 54.45 & 72.22 & 86.53 & +14.31 \\
J2T & Derived & 110 & 43.10 & 39.92 & 56.97 & 74.16 & +17.19 \\
J2T & Synthetic & 6 & 44.75 & 33.33 & 13.63 & 76.03 & +62.40 \\
T2J & Synthetic & 361 & 36.86 & 39.80 & 89.05 & 97.19 & +8.14 \\%
}

\newcommand{\ReviewSupportBaselinePairedRows}{%
TR & Claude-opus-5 & 101/45/38 & +10.55 & [+7.71, +17.86] \\
J2T & Gemini-3.5-flash & 63/39/14 & +19.53 & [+10.80, +28.60] \\
T2J & Claude-opus-5 & 75/276/10 & +8.14 & -- \\%
}

\newcommand{\ReviewSupportCoverageRows}{%
JC & 1063/1064 & 76.36 & 76.28 \\
SimJ & 855/856 & 79.47 & 79.41 \\%
}

\newcommand{\ReviewSupportGenerationConditionalRows}{%
Claude-opus-5 & 706 & 93.34 & 166 & 91.34 \\
Kimi-K3 & 729 & 90.14 & 143 & 85.24 \\
GPT-5.6-sol & 698 & 93.18 & 174 & 91.93 \\
Gemini-3.5-flash & 658 & 90.32 & 214 & 86.70 \\
Claude-opus-4.8 & 736 & 88.45 & 136 & 79.74 \\
GLM-5.3 & 696 & 95.16 & 176 & 92.56 \\
GPT-5.5 & 704 & 93.32 & 168 & 91.35 \\
Qwen3.8-max & 670 & 91.71 & 202 & 88.27 \\
GLM-5.2 & 608 & 88.87 & 264 & 84.77 \\
Kimi-2.6 & 590 & 90.92 & 282 & 87.85 \\
DeepSeek v4-Pro & 522 & 90.39 & 350 & 84.75 \\
MiniMax-M3 & 237 & 91.41 & 635 & 86.74 \\%
}

\newcommand{\ReviewSupportTopPairedRows}{%
Kimi-K3 & JDG & 941 & 941 & +1.19 & [+0.98, +1.40] \\
Kimi-K3 & PJFA & 872 & 872 & -1.43 & [-2.41, -0.43] \\
Kimi-K3 & QP & 714 & 714 & +0.36 & [-0.66, +1.39] \\
Kimi-K3 & JC & 1064 & 1064 & +0.88 & [-0.69, +2.43] \\
Kimi-K3 & TC & 161 & 161 & +0.00 & [-3.42, +3.73] \\
Kimi-K3 & PJF & 1386 & 874 & -0.97 & [-2.69, +0.69] \\
Kimi-K3 & TR & 184 & 18 & +1.53 & [-1.58, +2.40] \\
Kimi-K3 & SimT & 811 & 274 & -7.24 & [-13.43, +7.90] \\
Kimi-K3 & SimJ & 856 & 618 & +0.42 & [-1.09, +1.86] \\
Kimi-K3 & Dedup & 870 & 762 & +2.99 & [+1.65, +4.36] \\
Kimi-K3 & J2T & 116 & 24 & +0.94 & [-1.44, +2.70] \\
Kimi-K3 & T2J & 361 & 1 & +1.13 & -- \\
GPT-5.6-sol & JDG & 941 & 941 & +1.98 & [+1.80, +2.16] \\
GPT-5.6-sol & PJFA & 872 & 872 & -1.04 & [-2.13, +0.06] \\
GPT-5.6-sol & QP & 714 & 714 & +0.34 & [-0.61, +1.30] \\
GPT-5.6-sol & JC & 1064 & 1064 & -0.91 & [-2.47, +0.64] \\
GPT-5.6-sol & TC & 161 & 161 & -0.62 & [-4.35, +3.11] \\
GPT-5.6-sol & PJF & 1386 & 874 & -1.30 & [-3.19, +0.49] \\
GPT-5.6-sol & TR & 184 & 18 & +2.53 & [+1.87, +3.16] \\
GPT-5.6-sol & SimT & 811 & 274 & -0.43 & [-2.77, +4.14] \\
GPT-5.6-sol & SimJ & 856 & 618 & -0.33 & [-1.92, +1.16] \\
GPT-5.6-sol & Dedup & 870 & 762 & +0.92 & [-0.11, +1.95] \\
GPT-5.6-sol & J2T & 116 & 24 & +0.23 & [-2.06, +2.87] \\
GPT-5.6-sol & T2J & 361 & 1 & +0.03 & -- \\
Gemini-3.5-flash & JDG & 941 & 941 & +1.19 & [+1.03, +1.34] \\
Gemini-3.5-flash & PJFA & 872 & 872 & +2.86 & [+1.66, +4.10] \\
Gemini-3.5-flash & QP & 714 & 714 & +1.71 & [+0.49, +2.91] \\
Gemini-3.5-flash & JC & 1064 & 1064 & -0.52 & [-1.96, +0.88] \\
Gemini-3.5-flash & TC & 161 & 161 & +0.00 & [-2.80, +2.80] \\
Gemini-3.5-flash & PJF & 1386 & 874 & -2.20 & [-3.85, -0.59] \\
Gemini-3.5-flash & TR & 184 & 18 & +2.83 & [-0.50, +3.87] \\
Gemini-3.5-flash & SimT & 811 & 274 & -9.91 & [-14.38, -0.80] \\
Gemini-3.5-flash & SimJ & 856 & 618 & +0.19 & [-1.30, +1.74] \\
Gemini-3.5-flash & Dedup & 870 & 762 & +1.95 & [+0.72, +3.24] \\
Gemini-3.5-flash & J2T & 116 & 24 & -0.54 & [-3.71, +2.63] \\
Gemini-3.5-flash & T2J & 361 & 1 & +0.63 & -- \\%
}

\newcommand{\ExpertPjfConstructionRows}{%
Claude-opus-5 & 52.69 & 92.59 & 72.0 & 33.4 \\
Kimi-K3 & 54.20 & 94.44 & 74.3 & 34.1 \\
GPT-5.6-sol & 56.64 & 89.66 & 71.3 & 41.9 \\
Gemini-3.5-flash & 54.43 & 94.44 & 81.4 & 27.5 \\
Claude-opus-4.8 & 52.01 & 93.52 & 75.1 & 28.9 \\
GLM-5.3 & 54.75 & 92.75 & 74.9 & 34.6 \\
GPT-5.5 & 53.37 & 95.22 & 72.4 & 34.4 \\
Qwen3.8-max & 52.85 & 93.83 & 62.3 & 43.4 \\
GLM-5.2 & 55.58 & 93.21 & 78.5 & 32.7 \\
Kimi-2.6 & 55.27 & 93.83 & 65.3 & 45.3 \\
DeepSeek v4-Pro & 52.73 & 85.96 & 57.1 & 48.3 \\
MiniMax-M3 & 55.40 & 83.64 & 49.0 & 61.8 \\%
}

\newcommand{\ExpertPjfCvRows}{%
Resume & Resume + job & 0.661 & 0.002 & 61.01 \\
Resume & Resume only & 0.546 & 0.005 & 54.26 \\
Resume & Job only & 0.671 & 0.006 & 63.28 \\
Post & Resume + job & 0.651 & 0.005 & 60.30 \\
Post & Resume only & 0.539 & 0.008 & 52.62 \\
Post & Job only & 0.666 & 0.005 & 62.72 \\%
}

\newcommand{\ExpertGeneratorOverlapRows}{%
PJFA & 73.44 (9) & 76.17 (2) & 71.82 & 47.49--77.06 \\
TC & 97.86 (4) & 99.29 (1) & 94.00 & 85.00--99.29 \\
SimT & 53.89 (12) & 56.95 (10) & 67.82 & 55.25--75.30 \\
SimJ & 94.64 (7) & 99.07 (1) & 94.52 & 89.71--98.15 \\
Dedup & 95.67 (1) & 93.69 (3) & 85.50 & 67.56--94.88 \\%
}

\newcommand{\ExpertDifficultyStrataRows}{%
SimT & Same-category negatives: TNR & 167 & 58.08--98.20 \\
SimT & Cross-category negatives: TNR & 166 & 99.40--100.00 \\
SimJ & Same-category final negatives: TNR & 141 & 4.96--22.70 \\
SimJ & Cross-category negatives: TNR & 199 & 99.50--100.00 \\
Dedup & Masked identity positives: TPR & 250 & 93.60--100.00 \\
Dedup & Identity snapshots: TPR & 130 & 43.08--96.15 \\
Dedup & Matched-attribute negatives: TNR & 124 & 79.84--97.58 \\
QP & single & 325 & 79.82--87.95 \\
QP & multi constraint & 268 & 92.28--96.18 \\
QP & noisy & 121 & 77.27--88.55 \\%
}

\hypersetup{pdftitle={EHR-Bench: A Chinese Multi-Task Benchmark for Recruitment AI},
 pdfauthor={Ruijia Yang, Bing Xu, Zulong Chen, Zhipeng Xu, Jia Xu, Yipeng Yu, Ying Zhou, Zeyi Wen},
 pdfsubject={Preprint}, pdfkeywords={recruitment AI, multi-task benchmark, Chinese language}}
\title{\ehrbench: A Chinese Multi-Task Benchmark\\for Recruitment AI}

\author{%
\bfseries Ruijia Yang\textsuperscript{1}, Bing Xu\textsuperscript{2},
Zulong Chen\textsuperscript{2}, Zhipeng Xu\textsuperscript{2},\\
\bfseries Jia Xu\textsuperscript{3}, Yipeng Yu\textsuperscript{2},
Ying Zhou\textsuperscript{4,*}, Zeyi Wen\textsuperscript{1,*}\\[0.6em]
\normalfont\small\textsuperscript{1}The Hong Kong University of Science and Technology (Guangzhou)\\
\normalfont\small\textsuperscript{2}Alibaba Group\quad
\textsuperscript{3}Guangzhou University\\
\normalfont\small\textsuperscript{4}University of Electronic Science and Technology of China\\[0.3em]
\normalfont\small\textsuperscript{*}Corresponding authors.
}

\iclrfinalcopy

\begin{document}

\maketitle
\lhead{Preprint}

\begin{abstract}
Recruitment applications span diverse operations, from drafting job descriptions
and interpreting queries to organizing profiles, assessing fit, and recommending
candidates and jobs. Systematic LLM evaluation therefore requires task coverage that
reflects this range of business needs. We introduce \ehrbench, an open Chinese
benchmark comprising twelve tasks across six capability families and five recruitment
scenarios. The tasks cover generation, extraction, classification, retrieval,
semantic similarity, and reciprocal ranking within one evaluation framework.
An internal recruitment dataset guides task definitions, the occupation
taxonomy and sampling targets, output requirements, and scoring protocols. Following
these specifications, we construct 8,336 task instances from public materials and
controlled augmentation, with source-aware expert review and documented provenance.
We evaluate twelve LLMs through task-level, capability-level, and aggregate comparisons.
Closely ranked models exhibit distinct task strengths, and performance varies across
input sources, reference evidence, and candidate pools. Five shared models retain the
same observed Overall ordering on a private
operational reference, with variation in capability and task correspondence.
\ehrbench provides a reproducible foundation for systematic model assessment,
task-specific model selection, and further research on LLMs for recruitment.
The project repository is available at
\url{https://github.com/EhrBench/EHR-Bench}.
\end{abstract}

\section{Introduction}
\label{sec:introduction}

Recruitment applications use LLMs to draft job descriptions, interpret queries,
organize profiles, assess fit, and recommend candidates and jobs
\citep{otani2025nlphr,mashayekhi2024challenge}. These operations share resumes and job
descriptions but require different outputs: drafted content, structured attributes,
relationship judgments, and candidate orderings. Comparing models across these
operations requires a task framework that makes their distinct requirements explicit.

Existing resources address resume classification \citep{heakl2024resumeatlas}, skill
extraction \citep{zhang2022skillspan}, document parsing \citep{ling2025resumebench},
and person--job retrieval \citep{wang2026pjb}; WorkRB provides a shared framework for
work-domain tasks \citep{delange2026workrb}. A complementary need is a public Chinese
benchmark that connects diverse recruitment operations to enterprise task requirements.
We address this need with shared task specifications, documented inputs and reference
evidence, and task-specific scoring for aggregate and task-level model comparison.

\begin{figure}[t]
\centering\includegraphics[width=\textwidth]{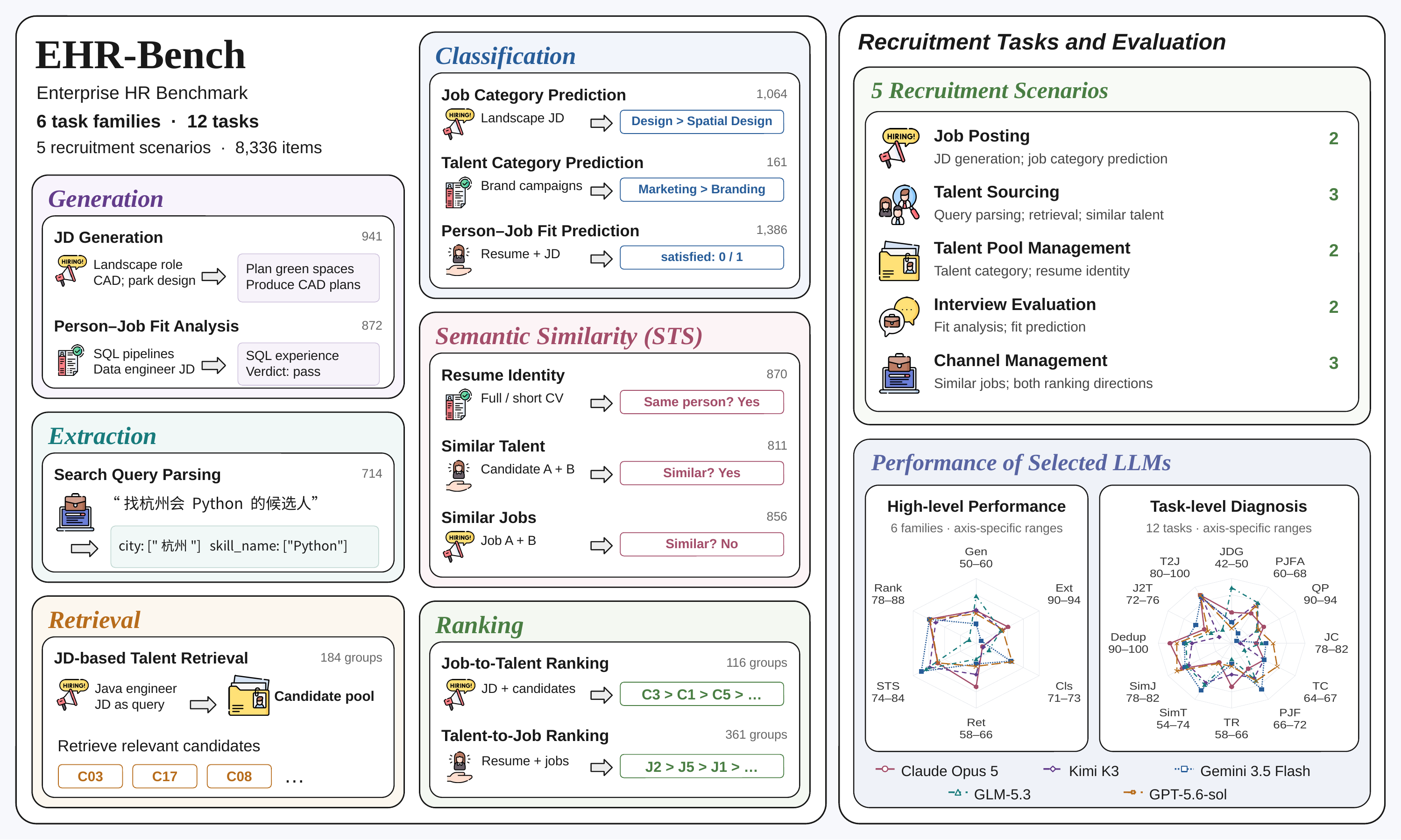}
\caption{Overview of \ehrbench: twelve tasks, six capability families, and five
 recruitment scenarios. Examples are illustrative; English text is translated.
 Radars show five leading model series, with scores scaled from center to rim over
 the range labeled on each axis.}
\label{fig:benchmark-overview}
\end{figure}

We introduce \ehrbench (Figure~\ref{fig:benchmark-overview}), an open Chinese benchmark
with \OpenRecordCount{} instances across twelve tasks, six capability families, and
five recruitment scenarios. An internal recruitment
dataset guides the task definitions, occupation taxonomy and sampling targets, output
schemas, and scoring. These specifications guide construction from public materials
and controlled augmentation, with source-aware expert review of reference labels.
The benchmark resources include data, provenance records, per-sample scoring exports, and evaluation tools
for reproducing scores and inspecting task-level differences.

Our twelve-LLM evaluation identifies distinct task advantages through capability
profiles and paired comparisons. The top four systems span 0.52 Overall points;
task-level results guide selection for specific recruitment operations. Source and
candidate-pool analyses further characterize performance across evaluation conditions.
Five shared LLMs retain the same observed
Overall order on the private operational reference that guided construction, while capability and task
correspondence varies.

We contribute (1) a framework that translates five recruitment scenarios into twelve
tasks and six capability families for cross-operation evaluation; (2) an open,
expert-reviewed resource whose construction is guided by an internal recruitment dataset,
with documented provenance and reusable evaluation tools; and (3) a twelve-LLM study
showing that models with similar Overall scores have distinct task-specific advantages
and that performance varies across input sources, reference evidence, and candidate
pools, informing task-specific model comparison and improvement.

\section{Related Work}
\label{sec:related-work}

\paragraph{Recruitment tasks and public benchmarks.}
Public resources cover complementary operations: resume classification
\citep{heakl2024resumeatlas}, skill extraction
\citep{zhang2022skillspan,li2026chineseskillspan}, multilingual resume parsing
\citep{ling2025resumebench}, and resume--JD questions and explanations
\citep{carrino2026jobresqa}. Candidate selection builds on requirement--experience
representations \citep{zhu2018pjfnn,qin2018apjfnn,li2020competence}, interaction evidence
\citep{bian2020coteaching}, and retrieval with hard-negative mining
\citep{yu2025confitv2}. PJB provides Chinese resume retrieval \citep{wang2026pjb},
while WorkRB supports ranking and multilabel classification
\citep{workrb2026software}. Table~\ref{tab:positioning} contrasts their evaluation
interfaces and candidate conditions. \ehrbench combines generation, structured
interpretation, relationship judgments, and candidate selection across five
recruitment scenarios, with enterprise-guided construction and shared public
protocols. Appendix~\ref{app:related} compares languages, task scope, output requirements,
and label evidence.

\begin{table}[!htbp]
\centering\fontsize{8}{9.6}\selectfont\setlength{\tabcolsep}{4pt}
\caption{Evaluation interfaces and candidate conditions. WorkRB coverage includes
 the ranking and classification interfaces in its accompanying toolkit \citep{workrb2026software}.}
\label{tab:positioning}
\begin{tabular}{@{}P{.25\textwidth}P{.32\textwidth}P{.38\textwidth}@{}}
\toprule
\rowcolor{EHRInk!5}\textbf{Resource} & \textbf{Language / output} & \textbf{Candidate conditions / scope} \\
\midrule
ResumeBench\newline\mbox{\citep{ling2025resumebench}} & Five languages; PDF to JSON & Schema-guided resume parsing \\
PJB\newline\mbox{\citep{wang2026pjb}} & Chinese; resume retrieval & 297 queries; 197,674-resume collection \\
WorkRB\newline\mbox{\citep{delange2026workrb}} & Multilingual; ranking and classification & Ontology- and corpus-based candidate spaces \\
\rowcolor{EHRGen!5}\textbf{EHR-Bench}\newline\textbf{(ours)} & \textbf{Chinese; text, spans, paths, relations, orderings} & \textbf{12 operations; 8--50-candidate retrieval and 5--10-candidate ranking pools} \\
\bottomrule
\end{tabular}
\end{table}

\paragraph{Multi-task evaluation and resource construction.}
HELM \citep{liang2023helm}, MTEB \citep{muennighoff2023mteb}, BEIR
\citep{thakur2021beir}, and LegalBench \citep{guha2023legalbench} motivate diverse task
coverage and capability-specific reporting. Synthetic CV resources
\citep{bruera2022syntheticcv} and JobSkape \citep{magron2024jobskape} provide construction
routes for employment text; data statements \citep{bender2018datastatements} and
datasheets \citep{gebru2021datasheets} organize provenance and intended use.
\ehrbench connects task coverage, construction, and evaluation: internal recruitment
operations guide task specifications and occupation sampling targets; public instances
retain documented reference evidence; and shared protocols support comparisons across
operations and between public and operational results for shared models.

\section{EHR-Bench: Design and Construction}
\label{sec:benchmark}

\subsection{Recruitment Scenarios and Task Framework}

Figure~\ref{fig:benchmark-overview} organizes recruitment operations at three levels.
Five scenarios provide the business contexts: job posting, talent sourcing, talent
pool management, interview evaluation, and channel management. Six capability
families group the abilities required across these contexts; twelve independently
evaluated tasks define inputs, outputs, references, and metrics
(Table~\ref{tab:task-inventory}; examples in Appendix~\ref{app:examples}).

Generation (Gen) covers JD drafting and fit analysis; Extraction (Ext) recovers
typed query spans; Classification (Cls) covers occupation assignment and binary fit
judgments. Retrieval (Ret) and Ranking (Rank) evaluate candidate selection, with
ranking in both directions. Semantic Textual Similarity (STS) covers relatedness
and identity. Shared resumes and jobs thus support complementary tests of content
production, structured interpretation, relationship judgments, and candidate decisions.

\begin{table}[!htbp]
\centering\fontsize{8}{9.8}\selectfont
\setlength{\tabcolsep}{3pt}\renewcommand{\arraystretch}{1.07}
\caption{Task definitions and abbreviations used throughout the paper. $n$ counts
 records ($\dagger$: query groups). $R$, $J$, $V$: ROUGE-L, rubric score, verdict
 accuracy; R@$k$: Recall@$k$.}
\label{tab:task-inventory}
\begin{tabular}{@{}lp{.265\textwidth}p{.285\textwidth}rp{.175\textwidth}@{}}
\toprule
\rowcolor{EHRInk!5}\textbf{Family} & \textbf{Task} & \textbf{Input $\rightarrow$ output} & \textbf{$n$} & \textbf{Main metric} \\
\midrule
\caplabel{EHRGen}{Gen} & Job description generation (JDG) & Job constraints $\rightarrow$ JD & 941 & $.6R+.4J$ \\
 & Person--job fit analysis (PJFA) & Resume + job $\rightarrow$ verdict, analysis & 872 & $.5V+.3R+.2J$ \\
\addlinespace[3pt]
\caplabel{EHRExt}{Ext} & Query parsing (QP) & Query $\rightarrow$ typed spans & 714 & Strict span-F1 \\
\addlinespace[3pt]
\caplabel{EHRCls}{Cls} & Job classification (JC) & Job text $\rightarrow$ category path & 1,064 & PathScore \\
 & Talent classification (TC) & Resume $\rightarrow$ category path & 161 & PathScore \\
 & Person--job fit (PJF) & Resume + job $\rightarrow$ binary fit & 1,386 & Balanced accuracy \\
\addlinespace[3pt]
\caplabel{EHRRet}{Ret} & Talent retrieval (TR) & Job + pool $\rightarrow$ candidate ordering & $184^\dagger$ & Mean R@3, R@10, MRR@10 \\
\addlinespace[3pt]
\caplabel{EHRSTS}{STS} & Talent similarity (SimT) & Two profiles $\rightarrow$ relatedness & 811 & Macro-F1 \\
 & Job similarity (SimJ) & Two jobs $\rightarrow$ relatedness & 856 & Macro-F1 \\
 & Resume deduplication (Dedup) & Two resumes $\rightarrow$ identity & 870 & Macro-F1 \\
\addlinespace[3pt]
\caplabel{EHRRank}{Rank} & Job-to-talent ranking (J2T) & Job + pool $\rightarrow$ talent ordering & $116^\dagger$ & nDCG@5 \\
 & Talent-to-job ranking (T2J) & Resume + pool $\rightarrow$ job ordering & $361^\dagger$ & nDCG@5 \\
\bottomrule
\end{tabular}
\end{table}

\subsection{Data Construction Guided by the Internal Dataset}

The internal dataset guides public task specifications
(Appendix~\ref{app:operational-mapping}): adopted JD drafts inform generation fields
and rubrics; HR assessments distinguish fit decisions from verdict-plus-analysis
outputs; annotated recruitment queries define entity schemas, while search and
recommendation interactions inform candidate interfaces. Occupation labels supply a shared three-level taxonomy with
eighteen first-level categories and sampling targets based on internal proportions.
Public and constructed examples implement these specifications with source-specific
references and expert review. Shared-model ordering is examined after construction
(Section~\ref{sec:analysis}).

External inputs combine public profiles, JDs, interaction records, and controlled
augmentation. Thirteen company job boards are accessed through the Greenhouse public API.

These postings supply 441 JDG and 339 JC instances: GPT-5.6-sol translates English postings,
humans proofread them, and cleaning removes company introductions and contact details.
Native Chinese postings follow a separate pathway. Claude-opus-5 and GPT-5.6-sol
each generate half of the persona-conditioned resume content, using occupation,
experience, city, industry, and education seeds; experts assign their task labels
(Appendix~\ref{app:resume-synthesis}). Resume-NER and a small Chinese adaptation of
ranking examples provide further augmentation. Each instance retains its source identifier.

Reference labels retain task- and source-specific evidence. Public JDs supply
generation references or receive occupation labels. Generated queries and candidate
entity outputs undergo review; classification uses model-assisted label proposals.
PJF combines 900 behavior-derived pairs, whose satisfaction labels undergo
mapping and applicability checks, with 486 added pairs labeled by domain experts
with initial suggestions from an internal fit model.
SimT and SimJ seed pairs from shared-job and
shared-candidate satisfaction, respectively, alongside occupational rules; experts
determine final similarity labels from the paired profiles or job descriptions.
Dedup identifies the same
person under controlled profile transformations.

Candidate tasks specify the anchor, eligible pool, and relevance evidence: TR combines
application-based and constructed pools; J2T uses satisfaction-derived relevance with a
small constructed supplement; T2J uses constructed relevance. Each ranking
group has one positive candidate; synthetic T2J fields named after behavior encode
constructed relevance rather than observed events. Appendix~\ref{app:construction}
details the operational labels and construction rules.

\subsection{Dataset Composition and Quality Assurance}

Each retrieval or ranking group counts once among the 8,336 benchmark instances.
TR has 184 groups (6,028 candidates), J2T 116 (847), and T2J 361 (2,888).

\begin{figure}[t]
\centering\includegraphics[width=\textwidth]{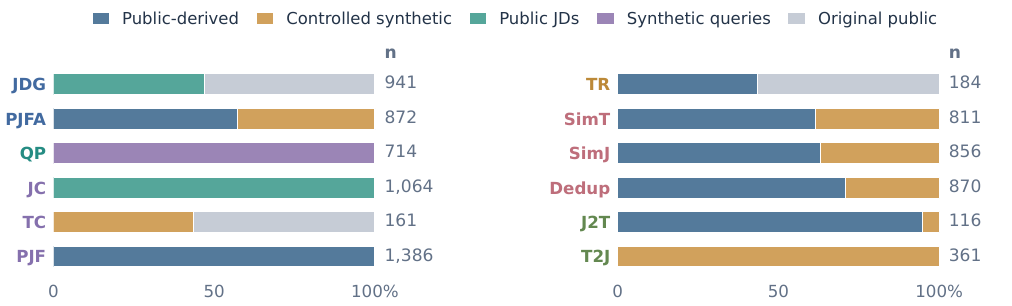}
\caption{Source composition by task. Colors indicate five categories of record provenance and construction;
 counts treat each retrieval or ranking group as one evaluation instance.}
\label{fig:composition}
\end{figure}

Figure~\ref{fig:composition} summarizes record sources and construction types by task.
Constructed records may reuse public text; label evidence is specified separately.
Within the enterprise reference's eighteen first-level
occupations, job classification covers all eighteen and talent classification
seventeen. TC sampling retains 161 of 735 records to target internal occupation
proportions. Section~\ref{sec:analysis}
compares these cohorts with the operational reference.

Schema, range, duplicate, and cross-field checks detect structural problems;
Qwen-plus flags potential semantic errors and disagreements. Three recruitment-domain experts
confirm labels against guidelines and original evidence, preserving inherited meanings
and adjudicating constructed relations. This source-aware workflow shares review
principles with CC-OCR v2 and UNIKIE-BENCH \citep{peng2026ccocrv2,ji2026unikie}.
Independent double annotation covers 760 records; the sample-size-weighted mean of
task-level Cohen's $\kappa$ is 0.86 for the 474 records from six discrete-label tasks.
A separate stratified audit confirms 395/400 records as correct.
Appendix~\ref{app:human-review} documents the review SOP, expert roles, and overlapping
quality-check subsets.

\section{Evaluation Protocol}
\label{sec:evaluation}

The protocol combines task-specific metrics, shared model settings, and explicit
output-handling rules to support reproducible model comparisons at task and
capability levels.

\subsection{Task Metrics and Capability Aggregation}

Task-specific metrics evaluate the outputs in Table~\ref{tab:task-inventory} on
0--100 scales; aggregation first averages within capabilities, then across them.
JDG combines character-level ROUGE-L ($R$; \citealp{lin2004rouge}) and a rubric
judge ($J$) as $0.6R+0.4J$: the components assess reference overlap and JD
quality, respectively. PJFA adds verdict accuracy ($V$), using $0.5V+0.3R+0.2J$;
all components use 0--100 scales. PJFA awards verdict
credit only for a reference-matching decision extracted by the fixed parser;
available analysis text is scored separately. The rubric-based
evaluation follows the LLM-as-a-judge approach \citep{zheng2023judge}.
Component scores distinguish verdict agreement, reference overlap, and rubric quality. Query parsing
uses corpus-level strict span micro-F1.
Occupation classification divides the correctly matched prefix length by the
reference depth. Similarity and identity tasks use macro-F1.

PJF uses balanced accuracy, $(\mathrm{TPR}+\mathrm{TNR})/2$, on parsed binary
predictions, with invalid predictions mapped to the negative label (0) before
calculation (Appendix~\ref{app:metrics}).
Retrieval averages Recall@3, Recall@10, and MRR@10
within each group, then averages groups. Reciprocal ranking uses nDCG@5
\citep{jarvelin2002ndcg} to score the single positive's position in each shortlist.

For tasks $T_c$ in category $c$, the reporting protocol is
\begin{equation}
S_c=|T_c|^{-1}\sum_{t\in T_c}s_t,\qquad
S_{\mathrm{Overall}}=\frac{1}{6}\sum_c S_c.
\end{equation}
Task weights $1/(6|T_c|)$ are $1/6$ in Extraction/Retrieval, $1/12$ in
Generation/Ranking, and $1/18$ in Classification/STS. Equal capability weights keep
each family's contribution independent of its task count; individual metrics retain
their task-specific meaning.

\subsection{Models and Evaluation Settings}

We evaluate twelve LLMs and tabulate task, capability, and Overall scores to two decimal places.
All 144 task scores are reproducible from per-sample exports
(Appendix~\ref{app:reproducibility}). Qwen-plus screens candidate labels and
Qwen-3.6-Flash applies common task-specific rubrics to generation outputs; neither is evaluated. Some evaluated
models assist content construction; experts adjudicate labels.

The evaluation runner uses fixed Chinese task prompts, plain-text JD generation,
and JSON for the other tasks. Temperature is zero except for Kimi-K3, which uses its endpoint default.
Thinking is disabled where configurable; GLM-5.3 uses low reasoning effort.
Occupation prompts contain the category tree and truncate input text at 3,500
characters. Ranking receives a candidate-dependent output budget.
Appendix~\ref{app:execution} records system settings, release identifiers, and runner
defaults; the artifact record identifies evaluation prompts, output parsers,
generation rubrics, and task scorers.

\subsection{Output Handling and Paired Comparisons}

Output handling and scored-set coverage follow task-specific rules. For generation,
classification, binary judgments, and query parsing, missing or empty raw outputs are omitted; nonempty
malformed outputs are handled by the task parser. Retrieval and ranking iterate over
all supplied groups. A ranking must contain every candidate exactly once after the
specified ID-repair rules, otherwise its group score is zero. Retrieval evaluates only
returned top-ten slots, with invalid or duplicate IDs receiving zero relevance.
All model--task sets are complete except two MiniMax-M3 refusals, one each in
JC and SimJ. Appendix~\ref{app:coverage} reports their identifiers and the effect
of counting them as failures.

Paired comparisons align scored sample IDs and use 10,000 bootstrap resamples of
connected components of observed entity IDs, preserving complete query groups
\citep{efron1993bootstrap}. We report paired 95\% bootstrap intervals for task-level
differences on the common scored subsets. Appendix~\ref{app:paired} lists dependence
units, fixed seeds, and single-component cases without intervals.

\section{Benchmark Results and Analysis}
\label{sec:experiments}

\subsection{Overall Performance and Task-Specific Strengths}

\begin{table}[H]
\centering\fontsize{8}{10}\selectfont\setlength{\tabcolsep}{4pt}\renewcommand{\arraystretch}{1.08}
\caption{Overall and capability scores. Bold / underline indicate the best /
 second-best distinct values, including ties. Overall averages the six capabilities.}
\label{tab:external-overall}
\begin{tabular}{lrrrrrrr}
\toprule
\rowcolor{EHRInk!5}\textbf{System} & \textbf{Overall} & \caphead{EHRGen}{Gen} & \caphead{EHRExt}{Ext} & \caphead{EHRCls}{Cls} & \caphead{EHRRet}{Ret} & \caphead{EHRSTS}{STS} & \caphead{EHRRank}{Rank} \\
\midrule
\TableOpenCategoryRows
\bottomrule
\end{tabular}
\end{table}

\begin{figure}[!htbp]
\centering\includegraphics[width=\textwidth]{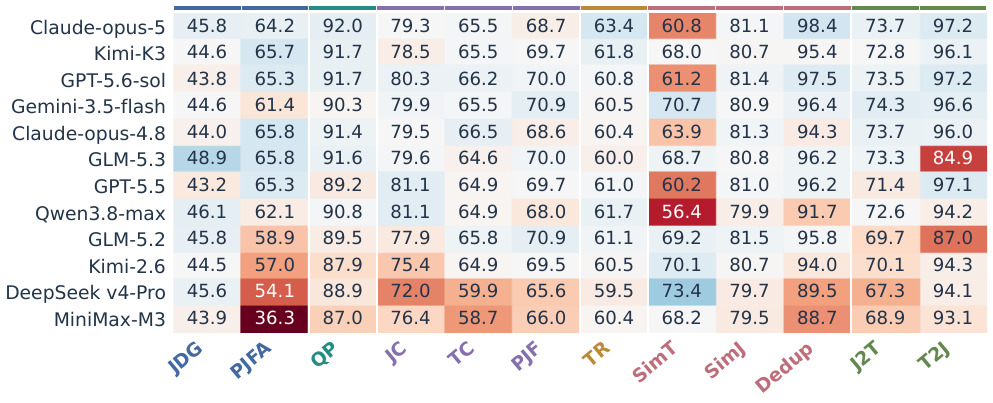}
\caption{Twelve-task scores in Overall order. Red/blue indicate scores below/above
 each task's median, clipped at $\pm15$ points; colors show within-task relative position,
 not cross-task difficulty.}
\label{fig:ranks}
\end{figure}

Overall scores range from 69.03 to 74.53 (Table~\ref{tab:external-overall}), with
a 0.52-point span across the top four. Across all twelve models, capability profiles
reveal distinct strengths: Claude-opus-5 leads Extraction, Retrieval, and Ranking; GPT-5.6-sol leads
Classification; Gemini-3.5-flash leads STS; and sixth-ranked GLM-5.3 leads Generation
(57.30). Figure~\ref{fig:ranks} extends this view to all twelve tasks, showing how
strengths vary across individual recruitment operations within the same Overall ordering.

Paired comparisons identify task-level advantages among closely ranked models.
Against Kimi-K3, Claude-opus-5
has a 1.19-point JDG advantage (paired 95\% interval $[0.98,1.40]$) and 2.99 on
Dedup ($[1.65,4.36]$), while Kimi-K3 leads PJFA by 1.43 ($[0.43,2.41]$).
These contrasts support task-specific model selection among closely ranked systems.
Omitting Retrieval or Ranking changes the leading system
(Appendix~\ref{app:top-paired}); Overall and task profiles support different selection needs.

\pagebreak[3]
\begin{samepage}
\subsection{Task-Level Diagnostics}

We examine taxonomy depth, generation components, and output validity to interpret task scores.

\paragraph{Occupation classification by taxonomy depth.}
Across all twelve systems, first-level job classification accuracy exceeds accuracy
through the second level by 15.13--19.83 percentage points, identifying fine-grained
occupation assignment as an improvement target
(Appendix~\ref{app:diagnostics}, Figure~\ref{fig:diagnostics}b).
\end{samepage}

\paragraph{Generation components.}
GLM-5.3 has a higher PJFA composite than Claude-opus-5 (65.75 versus 64.25) and
lower verdict accuracy (79.82\% versus 80.96\%). Higher ROUGE-L and rubric scores
account for the difference in composite scores
(Appendix~\ref{app:diagnostics}).

\paragraph{Output validity.}
MiniMax-M3's JSON parsing success rate on 872 PJFA outputs is 34.17\%, compared
with 97.82--100\% for the other systems, making JSON-format compliance a major
source of verdict-score loss. GLM-5.3 fails the
complete-permutation requirement on 43 of 361 T2J groups, which receive zero in
all-group nDCG@5
(Appendices~\ref{app:diagnostics} and~\ref{app:pjfa-json}).

\subsection{Performance across Evaluation Conditions}

We compare performance across candidate pools, reference-label sources, public and
synthetic inputs, and relation types under the same fixed task protocols.

\begin{table}[H]
\centering\fontsize{8}{10}\selectfont\setlength{\tabcolsep}{5pt}
\caption{Candidate-task baselines by source (0--100). Uniform: expected score
 over permutations. LLM: the best whole-task system, scored within each stratum.
 Synthetic J2T has six groups.}
\label{tab:candidate-baselines}
\begin{tabular}{llrrrr}\toprule
\rowcolor{EHRInk!5}Task & Source & $n$ & Uniform & TF-IDF & LLM \\\midrule
TR & Public interactions & 104 & 16.68 & 37.90 & 45.55 \\
TR & Derived pools & 80 & 53.34 & 72.22 & 86.53 \\
J2T & Derived & 110 & 43.10 & 56.97 & 74.16 \\
J2T & Synthetic & 6 & 44.75 & 13.63 & 76.03 \\
T2J & Synthetic & 361 & 36.86 & 89.05 & 97.19 \\
\bottomrule\end{tabular}
\end{table}

\paragraph{Candidate-pool conditions.}
Retrieval scores are lower on public-interaction TR pools than on derived pools.
The strongest whole-task LLM scores 45.55 and 86.53, respectively. On public-interaction
pools, character-bigram TF-IDF scores 37.90 and uniform orderings 16.68; both controls
also score higher on derived pools (Table~\ref{tab:candidate-baselines}). Synthetic
T2J reaches 89.05 with TF-IDF alone. These controls characterize the candidate
conditions underlying the LLM comparisons.

\begin{figure}[!htbp]
\centering\includegraphics[width=\textwidth]{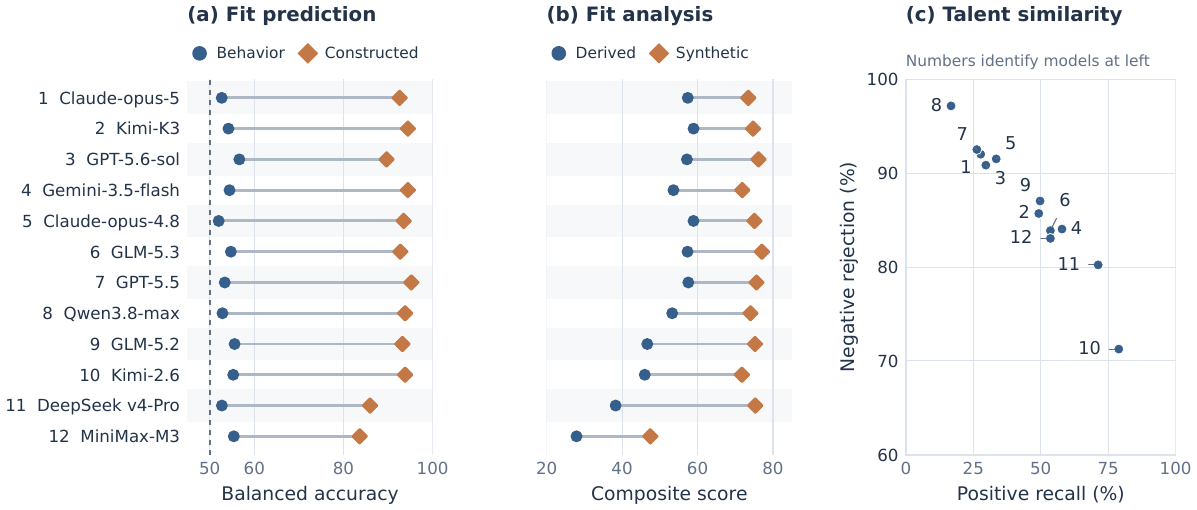}
\caption{Performance across evaluation conditions (0--100). (a) PJF: behavior-derived
 ($n=900$) and constructed ($486$); dashed line: chance at 50. (b) PJFA: interaction-derived ($500$) and synthetic
 ($372$). (c) SimT recall and negative rejection ($811$); numbers identify the systems
 listed at left.}
\label{fig:conditions}
\end{figure}

\paragraph{Reference-label sources.}
PJF performance differs across reference-label sources under the same fit instruction.
Across twelve LLMs, balanced accuracy is 52.01--56.64 on 900 behavior-derived
pairs versus 83.64--95.22 on 486 constructed pairs
(Figure~\ref{fig:conditions}a). Resume--job text cosine is inversely associated
with satisfaction (AUC 0.400). A supervised TF-IDF probe reaches 60.30--61.01
balanced accuracy with both texts under resume- or job-grouped cross-validation;
job-only features reach 62.72--63.28, versus 52.62--54.26 for resume-only features
(Appendix~\ref{app:pjf-construction}). These controls distinguish predictive job-side
associations from pairwise text overlap, complementing the comparison of model
agreement with the two reference sources.

\paragraph{Public and synthetic inputs.}
All systems score higher on synthetic PJFA and TC. Synthetic PJFA scores exceed
interaction-derived scores by 15.77--37.03 points; TC spans 38.46--44.51 on public
records versus 85.00--99.29 on synthetic records. The TC gap coincides with different input conditions: public profiles
have a median length of 219 characters, compared with 979 for synthetic resumes.
The gap identifies short public profiles as a challenging evaluation condition;
length, content, and label depth vary together across the two source strata
(Appendix~\ref{app:slices}).

\paragraph{Relation types and decision profiles.}
Models differ in which related profiles they recover: SimT positive recall ranges
from 16.75\% to 78.95\%, and negative rejection from 71.26\% to 97.18\%
(Figure~\ref{fig:conditions}c). On SimJ, all systems label 77.30--95.04\% of
the same-category pairs with final negative labels as similar, while rejecting 99.50--100\%
of cross-category negatives. The narrow aggregate range (79.47--81.47) therefore masks
low agreement on within-occupation negative pairs under the fixed similarity prompts
used for all twelve evaluated systems
(Appendix~\ref{app:operating-points}).

\section{Comparison with a Private Operational Reference}
\label{sec:analysis}

\subsection{Shared Specifications and Population Comparison}

After guiding public construction, the internal dataset provides an operational
reference under shared task definitions, taxonomy, schemas, and scoring.
We compare five shared LLMs: Kimi-K3, Qwen3.8-max, GLM-5.2, Kimi-2.6, and DeepSeek v4-Pro.

The classification cohorts provide a population comparison. External job/talent
classification covers 18/18 and 17/18 first-level occupations, respectively, with
total-variation distance 0.140 from the corresponding internal proportions.
Labels retain their task-specific evidence: recruiter acceptance, HR judgment,
public interactions, or constructed relevance. Appendix~\ref{app:population-definitions}
reports full occupation distributions together with input lengths, field coverage,
and candidate-pool characteristics (Table~\ref{tab:population}).

\subsection{Overall Agreement and Task-Level Variation}

Both collections give the same observed Overall ordering for the five shared models
(Spearman $\rho=1.00$, ten of ten pairs in the same direction).

Omitting each capability in turn from both collections and averaging the remaining
five gives cross-collection rank correlations of 0.90--1.00 at the reported score
precision (Appendix~\ref{app:operational_sensitivity}). The five-model aggregate
orderings remain closely aligned under all six omissions.

\begin{figure}[t]
\centering\includegraphics[width=\textwidth]{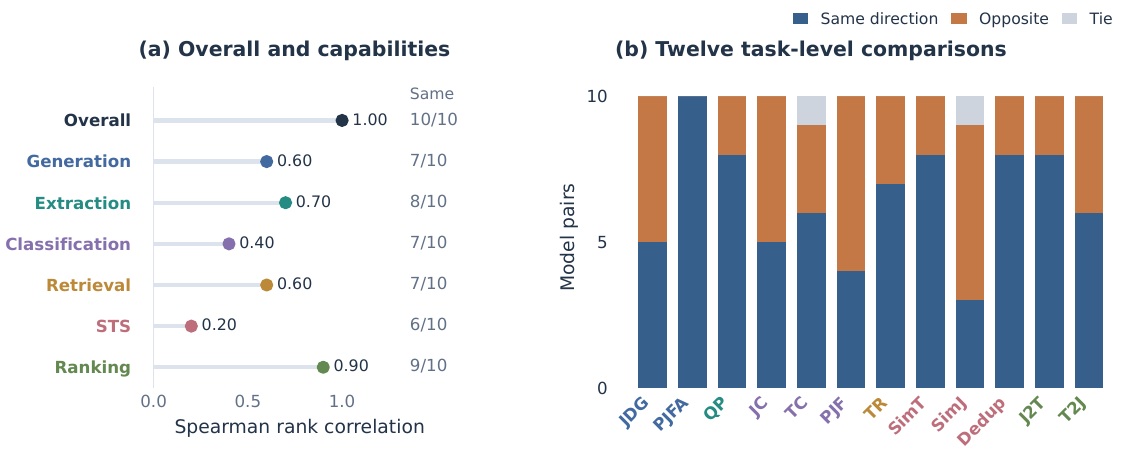}
\caption{Correspondence on five shared models. (a) Overall and capability Spearman
 correlations, with same-direction counts out of ten pairs. (b) Direction counts for
 each task at displayed precision; ties remain in the denominator of ten model pairs per task.}
\label{fig:operational}
\end{figure}

Capability correspondence ranges from $\rho=0.20$ for STS to $0.90$ for Ranking,
revealing variation beneath the shared Overall order (Figure~\ref{fig:operational}).

\begin{samepage}
Across twelve tasks, 78/120 pairwise comparisons agree in direction (65.0\%),
40 disagree, and two tie externally. Ties remain in the denominator. These task-level
comparisons locate differences in model ordering across the two collections
(Appendix~\ref{app:comparison}).\par
\end{samepage}

The production suite further illustrates task-specific strengths: its internal PJF
and T2J scores (74.30 and 65.98) exceed the best of the five LLMs (65.76 and 63.34),
while its Overall is 63.79.

For the five shared models, agreement in Overall ordering connects the public benchmark
to the operational reference; capability and task differences guide task-specific model selection.

\begin{samepage}
\section{Discussion and Limitations}
\label{sec:limitations}

The benchmark data, scorers, and per-sample scoring records support controlled comparisons
of models and prompting strategies across twelve recruitment tasks. Capability and
component profiles connect aggregate performance to specific operations, guiding model
selection and targeted improvements under shared inputs, references, and candidate pools.
\end{samepage}

Source analyses compare models under documented input and reference conditions.
Historical records supply interaction and satisfaction references, with experts
checking source-label mappings and applicability and judging newly constructed
relations. Content, length, and
reference evidence vary together across source strata. Candidate tasks assess selection
within supplied pools, with one positive per ranking group.

Experts assign task labels to model-generated resume content. Source-level comparisons
across five multi-source tasks show task-dependent strengths for the two resume
generators, with no consistent ranking advantage on synthetic records
(Appendix~\ref{app:generator-overlap}).

Scores measure task execution under the benchmark prompts, specified inference settings,
and fixed scoring protocols, including a shared generation rubric judge. These task-level
comparisons inform model screening; application-specific assessment addresses hiring
utility, exposure, and fairness (see Ethics statement;
\citealp{raghavan2020bias,kumar2023fairness,iso2025bias}).

\section{Conclusion}
\label{sec:conclusion}

\ehrbench establishes an open Chinese evaluation framework spanning twelve tasks,
six capability families, and five recruitment scenarios. Guided by an internal business
dataset, its construction translates enterprise task requirements into expert-reviewed
public data, documented references, and reproducible protocols. The twelve-LLM study
identifies task-specific strengths among closely ranked models and characterizes
performance across input, reference, and candidate conditions. The task
framework, open resource, and evaluation tools provide a shared basis for model comparison,
task-specific model selection, and research on models and prompting strategies for
recruitment.

\clearpage
\subsection*{Ethics statement}
The external collection combines public-source materials and constructed examples with
source cleaning, annotation, and review. Its intended use is research evaluation of
recruitment operations. Historical recruitment signals can reflect exposure and
selection patterns; deployment requires assessment in the relevant organizational and
fairness setting. Internal applicant and enterprise records remain in the controlled
environment, with aggregate statistics used for the comparison.

\subsection*{Reproducibility statement}
All 144 model--task scores are reproduced offline from per-sample predictions and
component statistics using the documented scoring formulas, within 0.011 points
on the 0--100 scale. Corpus, scorer, and result identifiers, coverage rules, and
reproduction commands appear in Appendix~\ref{app:reproducibility}. The saved artifacts support component, source, baseline, and paired analyses
using saved predictions, component statistics, and rubric scores. Aggregate private-reference tables
support reproduction of the correspondence statistics.

\subsection*{AI use statement}
AI models supported controlled generation and candidate-label preparation and checking.
For persona-conditioned resumes, Claude-opus-5 and GPT-5.6-sol generated equal shares
of the content and cross-reviewed each other's outputs. Qwen-plus assisted label-quality
screening, and an internal fit model supplied initial judgments for added PJF pairs.
A team of three domain experts reviewed final labels, checking inherited source labels
and confirming or revising newly assigned judgments.
Human reviewers also performed spot checks, independent double annotation, and anchor
review.
Qwen-3.6-Flash provided the common
rubric scores for generated outputs.
We distinguish data construction and annotation from output evaluation.
The authors led the research and manuscript preparation. Generative AI tools
assisted literature search, manuscript organization, paragraph drafting and
language editing, and analysis-code implementation. The authors take
responsibility for the final text, analyses, and conclusions.

\begingroup
\def\UrlBreaks{\do\/\do\_\do\-\do\?\do\&\do\=\do\#}
\bibliography{iclr2027_conference}

@article{heakl2024resumeatlas,
  title        = {{Resum\'{e}Atlas}: Revisiting Resume Classification with Large-Scale Datasets and Large Language Models},
  author       = {Heakl, Ahmed and Mohamed, Youssef and Mohamed, Noran and Elsharkawy, Aly and Zaky, Ahmed},
  journal      = {Procedia Computer Science},
  year         = {2024},
  url          = {https://www.sciencedirect.com/science/article/pii/S1877050924029909},
  volume       = {244},
  pages        = {158--165},
  doi          = {10.1016/j.procs.2024.10.189}
}

@inproceedings{ling2025resumebench,
  title        = {Beyond Human Labels: A Multi-Linguistic Auto-Generated Benchmark for Evaluating Large Language Models on Resume Parsing},
  author       = {Ling, Zijian and Zhang, Han and Cui, Jiahao and Wu, Zhequn and Sun, Xu and Li, Guohao and He, Xiangjian},
  booktitle    = {Proceedings of the 2025 Conference on Empirical Methods in Natural Language Processing},
  pages        = {31919--31945},
  year         = {2025},
  publisher    = {Association for Computational Linguistics},
  doi          = {10.18653/v1/2025.emnlp-main.1626},
  url          = {https://aclanthology.org/2025.emnlp-main.1626/}
}

@inproceedings{carrino2026jobresqa,
  title        = {{JobResQA}: Semi-Automatic Multilingual Benchmark Creation for {LLM} Machine Reading Comprehension on R\'{e}sum\'{e}s and Job Descriptions},
  author       = {Carrino, Casimiro Pio and Estrella, Paula and Zbib, Rabih and Escolano, Carlos and Fonollosa, Jos\'{e} A. R.},
  booktitle    = {Proceedings of the Fourth Workshop on the Role of Resources in the Age of Large Language Models (RESOURCEFUL 2026)},
  year         = {2026},
  pages        = {161--176},
  publisher    = {ELRA Language Resources Association (ELRA)},
  doi          = {10.63317/4pzwmqxt46xp},
  url          = {https://aclanthology.org/2026.resourceful-4.15/}
}

@article{wang2026pjb,
  title        = {{PJB}: A Reasoning-Aware Benchmark for Person-Job Retrieval},
  author       = {Wang, Guangzhi and Yang, Xiaohui and Li, Kai and He, Jiawen and Yang, Kai and Zhang, Ruixuan and Liu, Zhi},
  journal      = {arXiv preprint arXiv:2603.17386},
  year         = {2026},
  url          = {https://arxiv.org/abs/2603.17386}
}

@inproceedings{otani2025nlphr,
  title        = {Natural Language Processing for Human Resources: A Survey},
  author       = {Otani, Naoki and Bhutani, Nikita and Hruschka, Estevam},
  booktitle    = {Proceedings of the 2025 Conference of the Nations of the Americas Chapter of the Association for Computational Linguistics: Human Language Technologies (Volume 3: Industry Track)},
  pages        = {583--597},
  year         = {2025},
  publisher    = {Association for Computational Linguistics},
  doi          = {10.18653/v1/2025.naacl-industry.47},
  url          = {https://aclanthology.org/2025.naacl-industry.47/}
}

@inproceedings{muennighoff2023mteb,
  title        = {{MTEB}: Massive Text Embedding Benchmark},
  author       = {Muennighoff, Niklas and Tazi, Nouamane and Magne, Lo\"{i}c and Reimers, Nils},
  booktitle    = {Proceedings of the 17th Conference of the European Chapter of the Association for Computational Linguistics},
  pages        = {2014--2037},
  year         = {2023},
  publisher    = {Association for Computational Linguistics},
  doi          = {10.18653/v1/2023.eacl-main.148},
  url          = {https://aclanthology.org/2023.eacl-main.148/}
}

@inproceedings{thakur2021beir,
  title        = {{BEIR}: A Heterogeneous Benchmark for Zero-shot Evaluation of Information Retrieval Models},
  author       = {Thakur, Nandan and Reimers, Nils and R\"{u}ckl\'{e}, Andreas and Srivastava, Abhishek and Gurevych, Iryna},
  booktitle    = {Proceedings of the Neural Information Processing Systems Track on Datasets and Benchmarks},
  year         = {2021},
  url          = {https://datasets-benchmarks-proceedings.neurips.cc/paper/2021/hash/65b9eea6e1cc6bb9f0cd2a47751a186f-Abstract-round2.html},
  volume       = {1}
}

@article{zhu2018pjfnn,
  title        = {Person--Job Fit: Adapting the Right Talent for the Right Job with Joint Representation Learning},
  author       = {Zhu, Chen and Zhu, Hengshu and Xiong, Hui and Ma, Chao and Xie, Fang and Ding, Pengliang and Li, Pan},
  journal      = {ACM Transactions on Management Information Systems},
  volume       = {9},
  number       = {3},
  pages        = {1--17},
  year         = {2018},
  doi          = {10.1145/3234465},
  url          = {https://doi.org/10.1145/3234465}
}

@inproceedings{qin2018apjfnn,
  title        = {Enhancing Person--Job Fit for Talent Recruitment: An Ability-Aware Neural Network Approach},
  author       = {Qin, Chuan and Zhu, Hengshu and Xu, Tong and Zhu, Chen and Jiang, Liang and Chen, Enhong and Xiong, Hui},
  booktitle    = {Proceedings of the 41st International ACM SIGIR Conference on Research and Development in Information Retrieval},
  pages        = {25--34},
  year         = {2018},
  doi          = {10.1145/3209978.3210025},
  url          = {https://doi.org/10.1145/3209978.3210025}
}

@inproceedings{bian2020coteaching,
  title        = {Learning to Match Jobs with Resumes from Sparse Interaction Data Using Multi-View Co-Teaching Network},
  author       = {Bian, Shuqing and Chen, Xu and Zhao, Wayne Xin and Zhou, Kun and Hou, Yupeng and Song, Yang and Zhang, Tao and Wen, Ji-Rong},
  booktitle    = {Proceedings of the 29th ACM International Conference on Information and Knowledge Management},
  pages        = {65--74},
  year         = {2020},
  doi          = {10.1145/3340531.3411929},
  url          = {https://doi.org/10.1145/3340531.3411929}
}

@article{mashayekhi2024challenge,
  title        = {A Challenge-Based Survey of E-Recruitment Recommendation Systems},
  author       = {Mashayekhi, Yoosof and Li, Nan and Kang, Bo and Lijffijt, Jefrey and De Bie, Tijl},
  journal      = {ACM Computing Surveys},
  volume       = {56},
  number       = {10},
  pages        = {1--33},
  year         = {2024},
  doi          = {10.1145/3659942},
  url          = {https://doi.org/10.1145/3659942}
}

@inproceedings{raghavan2020bias,
  title        = {Mitigating Bias in Algorithmic Hiring: Evaluating Claims and Practices},
  author       = {Raghavan, Manish and Barocas, Solon and Kleinberg, Jon and Levy, Karen},
  booktitle    = {Proceedings of the 2020 Conference on Fairness, Accountability, and Transparency},
  pages        = {469--481},
  year         = {2020},
  doi          = {10.1145/3351095.3372828},
  url          = {https://doi.org/10.1145/3351095.3372828}
}

@article{kumar2023fairness,
  title        = {Fairness of Recommender Systems in the Recruitment Domain: An Analysis from Technical and Legal Perspectives},
  author       = {Kumar, Deepak and Grosz, Tessa and Rekabsaz, Navid and Greif, Elisabeth and Schedl, Markus},
  journal      = {Frontiers in Big Data},
  volume       = {6},
  pages        = {1245198},
  year         = {2023},
  doi          = {10.3389/fdata.2023.1245198},
  url          = {https://doi.org/10.3389/fdata.2023.1245198}
}

@article{liang2023helm,
  title        = {Holistic Evaluation of Language Models},
  author       = {Liang, Percy and Bommasani, Rishi and Lee, Tony and Tsipras, Dimitris and Soylu, Dilara and Yasunaga, Michihiro and Zhang, Yian and Narayanan, Deepak and Wu, Yuhuai and Kumar, Ananya and others},
  journal      = {Transactions on Machine Learning Research},
  year         = {2023},
  url          = {https://openreview.net/forum?id=iO4LZibEqW}
}

@article{bender2018datastatements,
  title        = {Data Statements for Natural Language Processing: Toward Mitigating System Bias and Enabling Better Science},
  author       = {Bender, Emily M. and Friedman, Batya},
  journal      = {Transactions of the Association for Computational Linguistics},
  volume       = {6},
  pages        = {587--604},
  year         = {2018},
  doi          = {10.1162/tacl_a_00041},
  url          = {https://aclanthology.org/Q18-1041/}
}

@article{gebru2021datasheets,
  title        = {Datasheets for Datasets},
  author       = {Gebru, Timnit and Morgenstern, Jamie and Vecchione, Briana and Vaughan, Jennifer Wortman and Wallach, Hanna and Daum\'{e} III, Hal and Crawford, Kate},
  journal      = {Communications of the ACM},
  volume       = {64},
  number       = {12},
  pages        = {86--92},
  year         = {2021},
  doi          = {10.1145/3458723},
  url          = {https://doi.org/10.1145/3458723}
}

@inproceedings{zheng2023judge,
  title        = {Judging {LLM}-as-a-Judge with {MT-Bench} and {Chatbot Arena}},
  author       = {Zheng, Lianmin and Chiang, Wei-Lin and Sheng, Ying and Zhuang, Siyuan and Wu, Zhanghao and Zhuang, Yonghao and Lin, Zi and Li, Zhuohan and Li, Dacheng and Xing, Eric P. and Zhang, Hao and Gonzalez, Joseph E. and Stoica, Ion},
  booktitle    = {Advances in Neural Information Processing Systems},
  volume       = {36},
  year         = {2023},
  doi          = {10.52202/075280-2020},
  pages        = {46595--46623},
  url          = {https://proceedings.neurips.cc/paper_files/paper/2023/hash/91f18a1287b398d378ef22505bf41832-Abstract-Datasets_and_Benchmarks.html}
}

@inproceedings{lin2004rouge,
  title        = {{ROUGE}: A Package for Automatic Evaluation of Summaries},
  author       = {Lin, Chin-Yew},
  booktitle    = {Text Summarization Branches Out},
  pages        = {74--81},
  year         = {2004},
  publisher    = {Association for Computational Linguistics},
  url          = {https://aclanthology.org/W04-1013/}
}

@article{peng2026ccocrv2,
  title        = {{CC-OCR V2}: Fine-Grained Attribution of {LMM} Failures in Real-World Visual Document Understanding},
  author       = {Peng, Chunyi and Xu, Zhipeng and Xiong, Yuqi and Chen, Zulong and Ji, Junhao and Liu, Qing and Liu, Zhenghao and Qin, Zubao and Zhao, Bing and Wan, Jianqiang and Tang, Jun and Yang, Zhibo and Bai, Shuai and Liu, Dayiheng and Sun, Maosong},
  journal      = {arXiv preprint arXiv:2605.03903},
  year         = {2026},
  doi          = {10.48550/arXiv.2605.03903},
  url          = {https://arxiv.org/abs/2605.03903}
}

@inproceedings{ji2026unikie,
  title        = {{UNIKIE-BENCH}: Benchmarking Large Multimodal Models for Key Information Extraction in Visual Documents},
  author       = {Ji, Yifan and Xu, Zhipeng and Liu, Zhenghao and Chen, Zulong and Zhang, Qian and Yang, Zhibo and Lin, Junyang and Gu, Yu and Yu, Ge and Sun, Maosong},
  booktitle    = {Proceedings of the 64th Annual Meeting of the Association for Computational Linguistics (Volume 1: Long Papers)},
  year         = {2026},
  url          = {https://aclanthology.org/2026.acl-long.287/},
  pages        = {6331--6352},
  doi          = {10.18653/v1/2026.acl-long.287}
}

@inproceedings{guha2023legalbench,
  title        = {{LegalBench}: A Collaboratively Built Benchmark for Measuring Legal Reasoning in Large Language Models},
  author       = {Guha, Neel and Nyarko, Julian and Ho, Daniel E. and R\'{e}, Christopher and Chilton, Adam and Narayana, Aditya and others},
  booktitle    = {Advances in Neural Information Processing Systems},
  volume       = {36},
  year         = {2023},
  doi          = {10.52202/075280-1915},
  pages        = {44123--44279},
  url          = {https://proceedings.neurips.cc/paper_files/paper/2023/hash/89e44582fd28ddfea1ea4dcb0ebbf4b0-Abstract-Datasets_and_Benchmarks.html}
}

@inproceedings{li2020competence,
  title        = {Competence-Level Prediction and Resume \& Job Description Matching Using Context-Aware Transformer Models},
  author       = {Li, Changmao and Fisher, Elaine and Thomas, Rebecca and Pittard, Steve and Hertzberg, Vicki and Choi, Jinho D.},
  booktitle    = {Proceedings of the 2020 Conference on Empirical Methods in Natural Language Processing},
  pages        = {8456--8466},
  year         = {2020},
  publisher    = {Association for Computational Linguistics},
  doi          = {10.18653/v1/2020.emnlp-main.679},
  url          = {https://aclanthology.org/2020.emnlp-main.679/}
}

@inproceedings{yu2025confitv2,
  title        = {{ConFit v2}: Improving Resume--Job Matching using Hypothetical Resume Embedding and Runner-Up Hard-Negative Mining},
  author       = {Yu, Xiao and Xu, Ruize and Xue, Chengyuan and Zhang, Jinzhong and Ma, Xu and Yu, Zhou},
  booktitle    = {Findings of the Association for Computational Linguistics: ACL 2025},
  pages        = {12775--12790},
  year         = {2025},
  publisher    = {Association for Computational Linguistics},
  doi          = {10.18653/v1/2025.findings-acl.661},
  url          = {https://aclanthology.org/2025.findings-acl.661/}
}

@article{fawcett2006roc,
  title        = {An Introduction to {ROC} Analysis},
  author       = {Fawcett, Tom},
  journal      = {Pattern Recognition Letters},
  volume       = {27},
  number       = {8},
  pages        = {861--874},
  year         = {2006},
  doi          = {10.1016/j.patrec.2005.10.010},
  url          = {https://doi.org/10.1016/j.patrec.2005.10.010}
}

@article{jarvelin2002ndcg,
  title        = {Cumulated Gain-Based Evaluation of {IR} Techniques},
  author       = {J\"{a}rvelin, Kalervo and Kek\"{a}l\"{a}inen, Jaana},
  journal      = {ACM Transactions on Information Systems},
  volume       = {20},
  number       = {4},
  pages        = {422--446},
  year         = {2002},
  doi          = {10.1145/582415.582418},
  url          = {https://doi.org/10.1145/582415.582418}
}

@book{efron1993bootstrap,
  title        = {An Introduction to the Bootstrap},
  author       = {Efron, Bradley and Tibshirani, Robert J.},
  publisher    = {Chapman and Hall},
  year         = {1993},
  edition      = {First},
  url          = {https://www.routledge.com/An-Introduction-to-the-Bootstrap/Efron-Tibshirani/p/book/9780412042317}
}

@inproceedings{magron2024jobskape,
  title        = {{JobSkape}: A Framework for Generating Synthetic Job Postings to Enhance Skill Matching},
  author       = {Magron, Antoine and Dai, Anna and Zhang, Mike and Montariol, Syrielle and Bosselut, Antoine},
  booktitle    = {Proceedings of the First Workshop on Natural Language Processing for Human Resources},
  pages        = {43--58},
  year         = {2024},
  publisher    = {Association for Computational Linguistics},
  doi          = {10.18653/v1/2024.nlp4hr-1.4},
  url          = {https://aclanthology.org/2024.nlp4hr-1.4/}
}

@inproceedings{bruera2022syntheticcv,
  title        = {Generating Realistic Synthetic Curricula Vitae for Machine Learning Applications under Differential Privacy},
  author       = {Bruera, Andrea and Alda, Francesco and Di Cerbo, Francesco},
  booktitle    = {Proceedings of the Workshop on Ethical and Legal Issues in Human Language Technologies and Multilingual De-Identification of Sensitive Data},
  pages        = {53--63},
  year         = {2022},
  publisher    = {European Language Resources Association},
  url          = {https://aclanthology.org/2022.legal-1.11/}
}

@article{delange2026workrb,
  title        = {{WorkRB}: A Community-Driven Evaluation Framework for {AI} in the Work Domain},
  author       = {De Lange, Matthias and Veys, Warre and Retyk, Federico and Deniz, Daniel and Jouanneau, Warren and Zhang, Mike and Bielinski, Aleksander and Jouffroy, Emma and Clobes, Nicole and Baranowska, Nina and Graus, David and Palyart, Marc and Zbib, Rabih and Gkatzia, Dimitra and Demeester, Thomas and De Bie, Tijl and Bogers, Toine and Decorte, Jens-Joris and Van Hautte, Jeroen},
  journal      = {arXiv preprint arXiv:2604.13055},
  year         = {2026},
  url          = {https://arxiv.org/abs/2604.13055}
}

@misc{workrb2026software,
  title        = {{WorkRB}: Work Research Benchmark},
  author       = {{TechWolf Research Team}},
  year         = {2026},
  url          = {https://github.com/techwolf-ai/workrb/tree/c417039140a2c60d87c302d4106554707b90b1a0},
  note         = {Version 0.6.0; software snapshot accessed September 21, 2026}
}

@article{cohen1960agreement,
  title        = {A Coefficient of Agreement for Nominal Scales},
  author       = {Cohen, Jacob},
  journal      = {Educational and Psychological Measurement},
  volume       = {20},
  number       = {1},
  pages        = {37--46},
  year         = {1960},
  doi          = {10.1177/001316446002000104},
  url          = {https://doi.org/10.1177/001316446002000104}
}

@inproceedings{iso2025bias,
  title        = {Evaluating Bias in {LLM}s for Job--Resume Matching: Gender, Race, and Education},
  author       = {Iso, Hayate and Pezeshkpour, Pouya and Bhutani, Nikita and Hruschka, Estevam},
  booktitle    = {Proceedings of the 2025 Conference of the Nations of the Americas Chapter of the Association for Computational Linguistics: Human Language Technologies (Volume 3: Industry Track)},
  pages        = {672--683},
  year         = {2025},
  publisher    = {Association for Computational Linguistics},
  doi          = {10.18653/v1/2025.naacl-industry.55},
  url          = {https://aclanthology.org/2025.naacl-industry.55/}
}

@inproceedings{sanchez2025unedaccess,
  title        = {Bilingual Evaluation of Language Models on General Knowledge in University Entrance Exams with Minimal Contamination},
  author       = {S\'{a}nchez Salido, Eva and Morante, Roser and Gonzalo, Julio and Marco, Guillermo and Carrillo-de-Albornoz, Jorge and Plaza, Laura and Amigo, Enrique and Garc\'{i}a, Andr\'{e}s Fernandez and Benito-Santos, Alejandro and Ghajari Espinosa, Adri\'{a}n and Fresno, Victor},
  booktitle    = {Proceedings of the 31st International Conference on Computational Linguistics},
  pages        = {6184--6200},
  year         = {2025},
  publisher    = {Association for Computational Linguistics},
  url          = {https://aclanthology.org/2025.coling-main.413/}
}

@inproceedings{zhang2022skillspan,
  title        = {{SkillSpan}: Hard and Soft Skill Extraction from {English} Job Postings},
  author       = {Zhang, Mike and Jensen, Kristian and Sonniks, Sif and Plank, Barbara},
  booktitle    = {Proceedings of the 2022 Conference of the North American Chapter of the Association for Computational Linguistics: Human Language Technologies},
  pages        = {4962--4984},
  year         = {2022},
  publisher    = {Association for Computational Linguistics},
  doi          = {10.18653/v1/2022.naacl-main.366},
  url          = {https://aclanthology.org/2022.naacl-main.366/}
}

@article{li2026chineseskillspan,
  title        = {{Chinese-SkillSpan}: A Span-Level Dataset for {ESCO}-Aligned Competency Extraction from {Chinese} Job Ads},
  author       = {Li, Guojing and Fu, Zichuan and Li, Junyi and Zhou, Wenxia and Wu, Xinyang and Yang, Jinning and Gao, Jingtong and Huang, Feng and Zhao, Xiangyu},
  journal      = {arXiv preprint arXiv:2604.23009},
  year         = {2026},
  url          = {https://arxiv.org/abs/2604.23009}
}
\bibliographystyle{iclr2027_conference}
\endgroup

\appendix
\clearpage
The appendices document construction and provenance (Appendix~\ref{app:construction}),
scoring and execution (Appendix~\ref{app:metrics}), full results and coverage
(Appendix~\ref{app:coverage}), component and paired diagnostics
(Appendices~\ref{app:diagnostics}--\ref{app:paired}), data conditions
(Appendix~\ref{app:slices}), and operational results (Appendix~\ref{app:comparison}),
followed by resource comparisons and the artifact record.

\raggedbottom
\setcounter{topnumber}{4}
\setcounter{totalnumber}{6}
\renewcommand{\floatpagefraction}{0.75}
\makeatletter
\setlength{\@fptop}{0pt}
\setlength{\@fpsep}{16pt}
\makeatother
\renewcommand{\arraystretch}{1.10}
\section{Task Interfaces and Construction Record}
\label{app:construction}

Each instance includes sample and task identifiers, language, input, and source
metadata; reference labels are stored separately. Pair tasks supply two objects,
and group tasks an anchor and eligible candidates. Generation returns a JD or a
verdict with analysis. Query parsing returns typed entities whose surface forms
are located in the query for exact span scoring. Occupation outputs are taxonomy
paths; binary outputs express the task-specific relation. Retrieval and reciprocal
ranking request complete orderings. Retrieval scores the first ten slots;
reciprocal ranking requires a complete permutation. Prompts and schemas will accompany
the public release.

\begin{table}[htbp]
\centering\fontsize{8}{10}\selectfont\setlength{\tabcolsep}{4pt}
\caption{Operational label evidence for relation and candidate tasks. Final labels
 follow task-specific review of the initial construction.}
\label{tab:label-operations}
\begin{tabular}{@{}P{.08\textwidth}P{.29\textwidth}P{.32\textwidth}P{.25\textwidth}@{}}
\toprule
\rowcolor{EHRInk!5}Task & External positive evidence & External negatives / construction & Internal counterpart \\
\midrule
PJF & Inherited satisfaction labels in 900 public pairs; expert judgments in 486 added pairs & Applied without satisfaction (215); browsed without application (207); expert-reviewed new pairings & Resume assessment pass / fail \\
TR & Application in public interaction pools; constructed relevance in derived pools & Other candidates within the supplied pool & Talent selected for assessment after JD-search exposure; sampled negatives \\
SimT & Shared-job satisfaction seeds; same second-level occupation seeds & Same-category or same-first-level negatives; cross-category random pairs; review sets final labels & Both candidates pass the final interview for the same job \\
SimJ & Shared-candidate satisfaction seeds; occupation-based pairs & Same-category and cross-category pairs; review sets final labels & Same candidate passes final interviews for both jobs \\
Dedup & Masked, lossy, or multiple snapshots of one identity & Distinct-person neighbors matched on occupation, education, or seniority; reviewed labels & Expert-labeled same-person / different-person pairs \\
J2T & Satisfaction-derived relevance; small constructed supplement & Remaining candidates in each single-positive pool & Recommended talent accepted by the user \\
T2J & Constructed compatibility, not an observed event & Remaining jobs in each single-positive pool & Recommended job applied to by the candidate \\
\bottomrule\end{tabular}
\end{table}

Browsing, application, and satisfaction are distinct public interaction signals.
The private reference includes recruiter acceptance, candidate applications, and assessment outcomes.
These event definitions specify the interpretation of each task's labels. Generation references, typed query annotations, and occupation
labels are described in the source and review subsections below.

SimT assesses similarity in professional background, career direction, and
seniority; SimJ assesses similarity in job responsibilities and skill requirements.
In the behavior-derived portion, satisfaction records linking two talents to the
same job, or the same talent to two jobs, seed candidate pairs. Shared interactions
and occupation categories guide sampling; HR experts determine the final similarity
labels from the paired profiles or job descriptions.

\paragraph{Layered annotation and review.}
We retain each candidate label's construction chain, distinguishing observed
interactions or public annotations from deterministic constructions and model-assisted
proposals. Schema, range, duplicate, and cross-field checks identify structural
problems. Qwen-plus flags potential semantic errors and disagreements for the expert
review described below; its suggestions do not determine the reference labels.
CC-OCR v2 describes initial annotation, additional review, disagreement
adjudication, and automatic consistency checks \citep{peng2026ccocrv2}.
UNIKIE-BENCH distinguishes converted from re-annotated data, using model-assisted
verification and human review \citep{ji2026unikie}. These approaches motivate
staged, source-aware verification; our final label decisions remain with the
recruitment-domain expert team.
Cross-model checks of generated resume content and rubric scoring of model outputs
serve separate purposes (Appendices~\ref{app:resume-synthesis} and~\ref{app:judge}).

\subsection{Human Annotation and Review}
\label{app:human-review}

\begin{figure}[!htbp]
\centering\includegraphics[width=\textwidth]{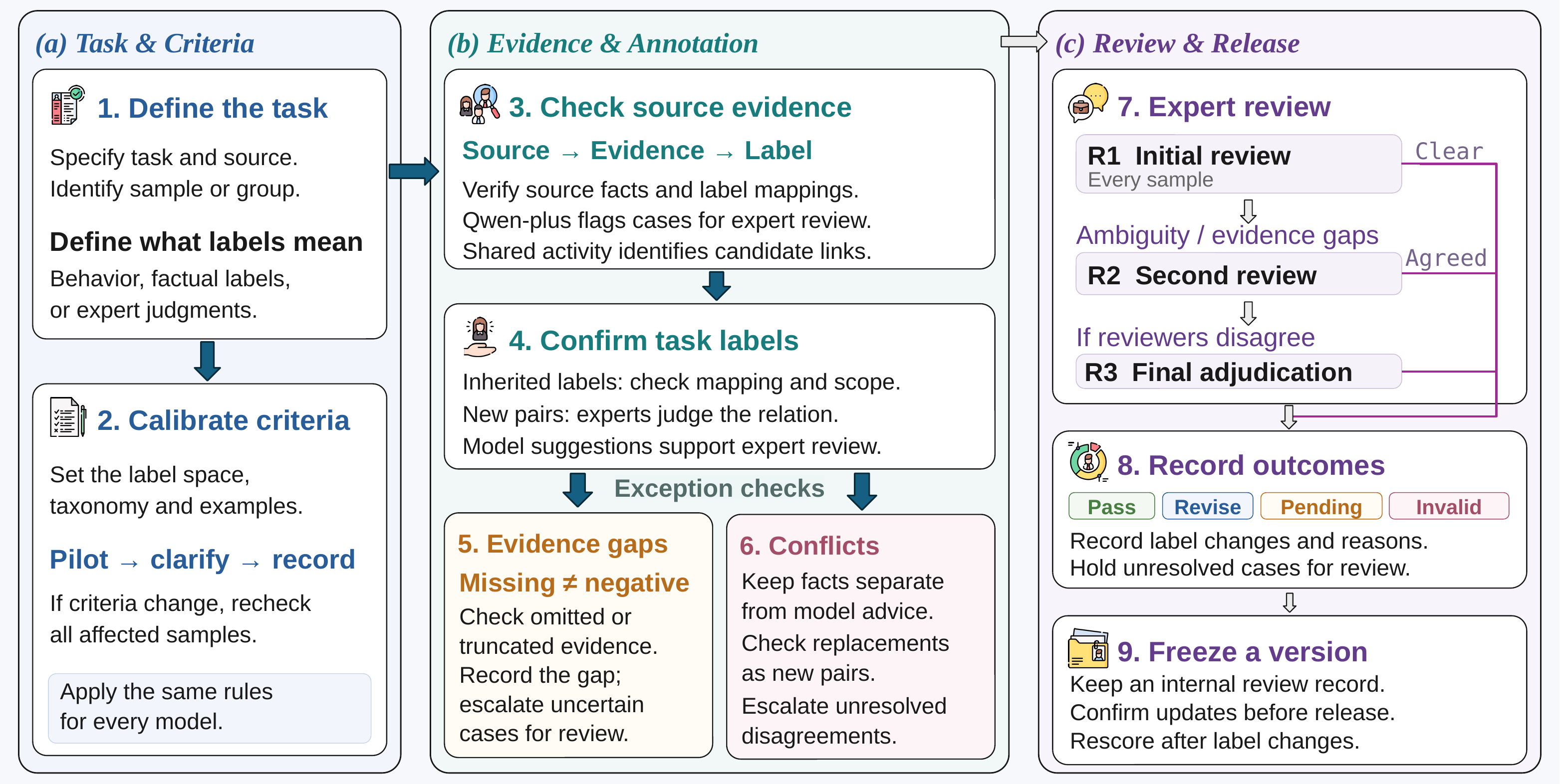}
\caption{Reference-label review SOP. R1 reviews all samples; R2 handles cases
 requiring further review; R3 adjudicates disagreements.}
\label{fig:annotation-sop}
\end{figure}

A team of three native Chinese domain experts with recruitment/HR and
NLP/annotation experience reviews reference labels using shared task guidelines and
original evidence. The standard operating procedure (SOP) in
Figure~\ref{fig:annotation-sop} separates source-label verification from judgment
of newly constructed relations. Qwen-plus supplies screening suggestions; experts
retain final authority over label acceptance and revision.

\paragraph{Source-dependent label handling.}
For the initial 900 public behavior-derived PJF pairs, labels pass from the
\texttt{satisfied} field in \texttt{table3\_action.csv}, through the original label,
to benchmark \texttt{reference.satisfied}. The satisfaction signal records HR
judgments based on the same resume and job information supplied to the models.
Review checks the mapping,
conversion, and applicability of that label. The behavioral references comprise
478 satisfaction positives, 215 application-without-satisfaction
negatives, and 207 browsing-without-application negatives.

The collection was subsequently expanded with 486 expert-labeled resume--job
pairs. An internal person--job fit model
provides an initial judgment as a review aid; domain experts determine the final
label from the new pair's evidence. The added pairs include 162 positives and 324
negatives. The full collection retains public-data-derived input provenance
(Appendix~\ref{app:pjf-construction}).

For SimT and SimJ, experts judge similarity from the paired resumes or job
descriptions. Shared interactions and occupation-based rules provide candidate
pairs; expert review confirms or revises their proposed labels.

\paragraph{Three-expert collaboration.}
The team first calibrates task scope, label spaces, and boundary examples. R1
conducts the initial expert review of every sample. Cases with unclear provenance,
missing evidence, or difficult judgments pass to R2 for additional review; R3
adjudicates disagreements using the original sources and task criteria. These roles
form an escalation process, not three independent annotations of every record.

The SOP distinguishes an unmet requirement from an unstated or undetermined one.
Insufficient evidence is flagged for review rather than automatically treated as a
negative judgment. Recorded behavior, textual evidence, and model suggestions are
kept distinct; verified mapping or conversion errors can be corrected without
reinterpreting the underlying event. Review outcomes are recorded as pass, revise,
pending, or invalid. Unresolved cases remain pending rather than being finalized
as definite labels. Confirmed changes retain their rationale and version history;
changes to task criteria require review of affected records under the updated rules.
The review procedure does not use an evaluated model's score to select labels.

Business-specific procedures and item-level expert-review records are not part of
the public release. The SOP documents the general workflow; aggregate quality
checks are reported below.

\paragraph{Independent quality checks.}
Additional quality checks comprise source-stratified spot checks on 1,250 records,
independent double annotation on 760, and review of 300 anchor examples. These
subsets overlap. Double annotation starts with two independent judgments; a senior
reviewer resolves disagreements. For PJF, both annotators independently assess fit
from the public resume and job inputs, without access to source labels.
These independent judgments are collected
separately from the escalation workflow above. Ambiguous cases inform guideline revisions.

Table~\ref{tab:human-review} summarizes agreement on the six discrete-label tasks.
The sample-size-weighted mean of task-level Cohen's $\kappa$
\citep{cohen1960agreement} is 0.86 across 474 records. The remaining double-annotated records cover generation,
extraction, and candidate tasks. A final stratified audit finds 395 of 400 records
correct (98.75\%), providing a separate check of the resulting labels.

\begin{table}[htbp]
\centering\small\setlength{\tabcolsep}{12pt}
\caption{Independent annotation agreement before adjudication.}
\label{tab:human-review}
\begin{tabular}{lrr}\toprule
\rowcolor{EHRInk!5}Task & Double-annotated records & Cohen's $\kappa$ \\\midrule
JC & 98 & 0.89 \\
TC & 20 & 0.83 \\
PJF & 125 & 0.86 \\
SimT & 74 & 0.80 \\
SimJ & 78 & 0.82 \\
Dedup & 79 & 0.92 \\
\bottomrule\end{tabular}
\end{table}
\FloatBarrier

\subsection{Operational Basis of Task Design}
\label{app:operational-mapping}

Table~\ref{tab:operational-mapping} connects the operations represented in the
enterprise reference to the twelve external tasks. Shared specifications guide
external construction using public and constructed materials, with task-specific
labels established through the evidence and review processes documented in this appendix.

\begin{table}[!htbp]
\centering\fontsize{8.5}{10.5}\selectfont
\setlength{\tabcolsep}{4pt}\renewcommand{\arraystretch}{1.12}
\caption{Business operations, corresponding tasks, and external construction guidance.}
\label{tab:operational-mapping}
\begin{tabular}{@{}P{.37\textwidth}P{.12\textwidth}P{.43\textwidth}@{}}
\toprule
\rowcolor{EHRInk!5}\textbf{Operational evidence} & \textbf{Tasks} & \textbf{External design and construction guidance} \\
\midrule
Vacancy drafting: JD outputs adopted by platform users.
& JDG
& Required job fields, the JD-generation operation, and a shared quality rubric. \\
\addlinespace[4pt]
Candidate assessment: screening pass/reject decisions and HR-written interview assessments.
& PJF, PJFA
& Distinct fit-decision and verdict-plus-analysis outputs; shared analysis rubric. \\
\addlinespace[4pt]
Search understanding: recruitment queries with annotated entity spans.
& QP
& Typed entities, span-based extraction, and the query-output schema. \\
\addlinespace[4pt]
Profile organization: operationally maintained occupation labels for jobs and talents.
& JC, TC
& A shared hierarchical occupation space, category-path outputs, and occupation-based sampling targets. \\
\addlinespace[4pt]
Talent discovery: candidates scheduled for assessment after JD-search exposure.
& TR
& Job anchors, eligible candidate pools, and relevance-based retrieval evaluation. \\
\addlinespace[4pt]
Related-profile discovery: talents passing final interviews for the same job, or jobs for which the same talent passes.
& SimT, SimJ
& Recruitment-domain similarity criteria; shared interactions seed pairs for review, alongside random and occupation-related constructions. \\
\addlinespace[4pt]
Identity resolution: expert-labeled resume versions and different-person pairs.
& Dedup
& Same-person identity judgments, controlled profile variations, and negative-pair construction. \\
\addlinespace[4pt]
Reciprocal recommendation: accepted talent recommendations and jobs applied to by talents.
& J2T, T2J
& Direction-specific recommendation inputs, candidate-list interfaces, and relevance-based ordering. \\
\bottomrule
\end{tabular}
\end{table}

\subsection{Source Resources and Transformations}
\label{app:sources}

The source tags in Figure~\ref{fig:composition} describe record sources and construction types.
Table~\ref{tab:source-resources} maps source materials to their uses.
Source text, task transformation, and label evidence are
separate fields: an interaction label and a constructed relation can use similar
profile text while relying on different forms of reference-label evidence.

\begin{table}[htbp]
\centering\small
\caption{Record-level source counts, including query groups. Percentages use
 8,336 records and may not sum to 100 through rounding.}
\begin{tabular}{lrr}\toprule\rowcolor{EHRInk!5}Source tag & Records & Percent \\
\midrule\TableSourceRows\bottomrule\end{tabular}
\end{table}

\begin{table}[!htbp]
\centering\small
\caption{Source resources, construction uses, and label evidence.}
\label{tab:source-resources}
\begin{tabular}{P{.28\textwidth}P{.29\textwidth}P{.34\textwidth}}
\toprule\rowcolor{EHRInk!5}Resource / material & Construction use & Label evidence \\
\midrule
\href{https://tianchi.aliyun.com/dataset/44080}{Tianchi--Zhaopin competition data} & Recruitment text and interaction-derived pairs or pools & Observed interaction fields for the original relation; task rules or reviewed judgments for constructed relations \\
\addlinespace[4pt]
\href{https://www.modelscope.cn/datasets/iic/resume_ner}{\texttt{iic/resume\_ner}} & Resume text for talent classification and identity examples & Occupation labeling, same-person transformations, and negative pairs \\
\addlinespace[4pt]
\href{https://github.com/GunaPalanivel/Avera}{Avera} & Small Chinese adaptation used in reciprocal-ranking augmentation & Task-specific candidate relevance \\
\addlinespace[4pt]
Greenhouse company job boards & Cleaned JD references and occupation classification & Reference text or reviewed category labels \\
\addlinespace[4pt]
Controlled synthetic materials & Persona-based resumes and other task-specific constructions & Expert labels for persona-generated resumes; task-specific rules and review elsewhere \\
\bottomrule
\end{tabular}
\end{table}

\paragraph{Original resources and license references.}
The recruitment profiles, job descriptions, and interaction records originate from the
\href{https://tianchi.aliyun.com/competition/entrance/231728/introduction}{Tianchi--Zhaopin person--job matching competition}
and its \href{https://tianchi.aliyun.com/dataset/44080}{preliminary dataset}.
The competition-data license reference is
\href{https://creativecommons.org/licenses/by-sa/4.0/}{CC BY-SA 4.0}.
The \texttt{iic/resume\_ner} resource is distributed by ModelScope's \texttt{iic}
account; its copyright notice specifies
\href{https://creativecommons.org/licenses/by/4.0/}{CC BY 4.0}.
These references apply to the corresponding source materials and their derivatives.

\paragraph{Task preparation and review.}
DeepSeek v4-Pro assists the initial cleaning and interface standardization of public
JDs. Task records follow the interfaces and occupation taxonomy in
Section~\ref{sec:benchmark}. Query references record the entities
expressed in each query; relation labels use interaction evidence, construction
rules, and expert judgments as appropriate to each task. Qwen-plus assists screening;
experts verify source-label mappings and confirm labels for newly constructed
examples (Appendix~\ref{app:human-review}).

The resource contains 8,336 records, counting each query group once. Expanding
groups into candidate rows gives 17,438 rows, a separate measure of candidate exposure.

\subsection{Public Job-Description Collection}
\label{app:jd-collection}

The additional JD collection comprises over 1,000 postings from thirteen company
careers boards, accessed through the
\href{https://docs.greenhouse.io/job-board.html}{Greenhouse Job Board API}.
The public JSON endpoint includes the posting text when \texttt{content=true}.
Table~\ref{tab:jd-boards} records the source boards; the collection count refers to
retrieved postings, while Table~\ref{tab:task-inventory} counts final task instances.
English posting content is translated into Chinese by GPT-5.6-sol and manually
proofread. Cleaning also removes company introductions and contact details before
task construction. The final collection contains 441 translated JDG references and
339 translated JC inputs. The other 725 JC inputs are native Chinese public
postings; these share the public-JD provenance tag but not the translation pathway.
The shared endpoint pattern is:

\begin{center}\footnotesize
\nolinkurl{https://boards-api.greenhouse.io/v1/boards/{board}/jobs?content=true}
\end{center}

\begin{table}[!htbp]
\centering\small\setlength{\tabcolsep}{7pt}
\caption{Company boards used for public JD collection; tokens link to JSON endpoints.}
\label{tab:jd-boards}
\begin{tabular}{@{}ll@{\hspace{2em}}ll@{}}
\toprule\rowcolor{EHRInk!5}Company & Board token & Company & Board token \\
\midrule
Stripe & \href{https://boards-api.greenhouse.io/v1/boards/stripe/jobs?content=true}{\texttt{stripe}} & Planet Labs & \href{https://boards-api.greenhouse.io/v1/boards/planetlabs/jobs?content=true}{\texttt{planetlabs}} \\
Anthropic & \href{https://boards-api.greenhouse.io/v1/boards/anthropic/jobs?content=true}{\texttt{anthropic}} & Esri & \href{https://boards-api.greenhouse.io/v1/boards/esri/jobs?content=true}{\texttt{esri}} \\
Coinbase & \href{https://boards-api.greenhouse.io/v1/boards/coinbase/jobs?content=true}{\texttt{coinbase}} & Gotion & \href{https://boards-api.greenhouse.io/v1/boards/gotion/jobs?content=true}{\texttt{gotion}} \\
Cloudflare & \href{https://boards-api.greenhouse.io/v1/boards/cloudflare/jobs?content=true}{\texttt{cloudflare}} & Flexport & \href{https://boards-api.greenhouse.io/v1/boards/flexport/jobs?content=true}{\texttt{flexport}} \\
Roblox & \href{https://boards-api.greenhouse.io/v1/boards/roblox/jobs?content=true}{\texttt{roblox}} & project44 & \href{https://boards-api.greenhouse.io/v1/boards/project44/jobs?content=true}{\texttt{project44}} \\
Taboola & \href{https://boards-api.greenhouse.io/v1/boards/taboola/jobs?content=true}{\texttt{taboola}} & ShipMonk & \href{https://boards-api.greenhouse.io/v1/boards/shipmonk/jobs?content=true}{\texttt{shipmonk}} \\
Wolt & \href{https://boards-api.greenhouse.io/v1/boards/wolt/jobs?content=true}{\texttt{wolt}} & & \\
\bottomrule
\end{tabular}
\end{table}

\subsection{Persona-Conditioned Resume Synthesis}
\label{app:resume-synthesis}

\paragraph{Persona design.}
Seeds combine a public three-level occupation tree (18 first-level groups) with
experience bands (new graduate, 1--3, 3--5, 5--10, and 10+ years), city, industry,
and education. An occupation-specific skill dictionary constrains skill selection.
These public persona categories specify the content-generation conditions.

\paragraph{Content generation and expert labels.}
Claude-opus-5 and GPT-5.6-sol each generate half of the synthetic resume content in
this stage. The structured JSON includes basic information, education, work experience
with quantified achievements, projects, and skills. Generation targets 800--2,000
Chinese characters and two to four work-experience sections. Human experts assign
the associated task labels. The two generators also appear in the evaluated system
set; their content-generation role is recorded separately from expert labeling and
the common rubric judge used for output scoring.

\paragraph{Quality control.}
Rule checks cover schema completeness, timeline consistency, and required skills.
Each generator's outputs are reviewed by the other model: Claude reviews GPT-generated
resumes and GPT reviews Claude-generated resumes. Realism, internal consistency,
and occupation fit are scored on a 1--5 scale, with a quality threshold of 4;
outputs below the threshold are regenerated with feedback. MinHash deduplication
completes the pipeline. Generator and evaluator comparisons appear in
Appendix~\ref{app:generator-overlap}.

\subsection{Illustrative Task Instances}
\label{app:examples}

Table~\ref{tab:task-examples} uses author-written, English-language illustrations
of the Chinese task interfaces. The snippets are schematic inputs and target forms,
not dataset records or model predictions; identifiers and people are fictitious.
For a resume and job, a fit decision returns a binary label, fit analysis pairs a
verdict with an explanation, and candidate selection introduces a supplied comparison
pool. The same objects therefore support distinct evaluation operations.

\small
\begin{longtable}{P{.12\textwidth}P{.36\textwidth}P{.43\textwidth}}
\caption{Illustrative task inputs and outputs; field names follow evaluation prompts.}
\label{tab:task-examples}\\
\toprule\rowcolor{EHRInk!5}Task & Illustrative input & Target form and task meaning \\
\midrule\endfirsthead
\toprule\rowcolor{EHRInk!5}Task & Illustrative input & Target form and task meaning \\
\midrule\endhead
JDG & Data engineer; SQL; maintain data pipelines. & A reference job description stating the role's duties and requirements. \\
\addlinespace[5pt]
PJFA & Resume mentions SQL pipelines; job requires SQL and pipeline maintenance. & \texttt{verdict: pass}; \texttt{analysis\_text}: explanation of the stated experience and requirements. \\
\addlinespace[5pt]
QP & ``Find SQL engineers in City A.'' & \texttt{skill\_name: [SQL]}, \texttt{jobname: [engineers]}, \texttt{city: [City A]}; surface strings are mapped back to exact query spans. \\
\addlinespace[5pt]
JC & A job description focused on data-pipeline development. & \texttt{cate\_labels}: an ordered occupation path selected from the supplied taxonomy. \\
\addlinespace[5pt]
TC & A resume describing data-pipeline development experience. & \texttt{cate\_labels}: the talent's occupation path in that taxonomy, rather than a label for a job vacancy. \\
\addlinespace[5pt]
PJF & A supplied resume--job pair. & \texttt{satisfied: 1} or \texttt{0}; the reference relation follows the pair's behavioral or constructed label evidence. \\
\addlinespace[5pt]
TR & A job and candidates with IDs \texttt{r1}, \texttt{r2}, \ldots & \texttt{ranking: [r2, r1, \ldots]}; the prompt requests all IDs, while relevance is evaluated in the first ten slots. \\
\addlinespace[5pt]
SimT & Profile A: mid-level Java backend developer. Profile B: Java engineer with comparable backend experience. & \texttt{is\_similar: 1}; similar professional direction and seniority. \\
\addlinespace[5pt]
SimJ & Job A: build Java backend services. Job B: develop Java APIs and maintain backend services. & \texttt{is\_similar: 1}; similar responsibilities and skill requirements. \\
\addlinespace[5pt]
Dedup & Two versions of a fictitious person's resume, with one shortened. & \texttt{is\_same\_person: 1}; the target is identity under the specified transformations. \\
\addlinespace[5pt]
J2T & A job and a five-person eligible pool, \texttt{r1}--\texttt{r5}. & \texttt{ranking: [r3,r1,r5,r2,r4]}; all five IDs must occur exactly once, and the labeled positive determines nDCG@5. \\
\addlinespace[5pt]
T2J & A resume and an eight-job pool, \texttt{j1}--\texttt{j8}. & \texttt{ranking:} \texttt{[j2,j5,j1,j3,} \texttt{j4,j6,j7,j8]}; a full permutation in the reciprocal direction. \\
\bottomrule
\end{longtable}
\normalsize

\section{Metric and Execution Details}
\label{app:metrics}

The benchmark first averages tasks within each capability, then averages the six
capabilities. Each Extraction or Retrieval task contributes $1/6$ of Overall;
each Generation or Ranking task $1/12$; each Classification or STS task $1/18$.
The weights depend only on the task design, so adding a system does not change
an existing system's score. Ranks and score ranges are descriptive views of the
reported results. Scores are not standardized or weighted by observed dispersion.

All metric descriptions here refer to the scoring implementation. ROUGE-L is
computed over Chinese characters. JD generation uses $0.6R+0.4J$; PJF analysis uses
$0.5V+0.3R+0.2J$, with each component on a 0--100 scale. The judge is
Qwen-3.6-Flash under the task-specific rubric. Occupation PathScore is the correctly
predicted prefix length divided by the reference path depth, with the first level
serving as the gate for subsequent levels. PJF balanced accuracy consumes parsed
binary outputs. The scorer names this metric AUC: for binary scores, ROC-AUC
\citep{fawcett2006roc} equals balanced accuracy when both classes are present,
using average ranks for ties. Invalid predictions map to the negative label (0) before this calculation;
macro-F1 retains an invalid prediction as an error for its true class.

Retrieval evaluates the explicitly returned first ten slots. Duplicate and invalid
identifiers occupy a slot with zero relevance. Recall uses the number of positives in
the full supplied pool as its denominator, and MRR is truncated at those ten slots.
Reciprocal ranking requires a complete permutation of the supplied candidates; a
missing, duplicate, or unrepairable identifier makes the group score zero. The shared
ID-repair routine accepts exact IDs, unique valid prefixes, or a unique fuzzy match
above similarity 0.9 with a margin greater than 0.05. No missing candidate is appended
from the input order. The two directions use nDCG@5; MRR and other cutoffs are diagnostic
outputs.

Missing/empty raw predictions are skipped by generation, query, occupation, and binary
scorers. A nonempty malformed query produces an empty entity set; malformed occupation
output yields no matched prefix. Generation uses the raw text fallback for text
comparison, and PJFA verdict comparison applies to the parsed field. Ranking and
retrieval iterate over the full task records, retaining empty-output groups as zero.
These rules define the scored-set denominators.

\subsection{Execution Interfaces and System Identifiers}
\label{app:execution}

Table~\ref{tab:execution-settings} summarizes decoding settings and the evaluation
runner's input and output budgets. Temperature is 0.0 for all
systems except Kimi-K3, whose endpoint uses its default temperature. Thinking is
disabled where configurable; GLM-5.3 uses \texttt{reasoning\_effort=low}.
DeepSeek v4-Pro was accessed through the vendor's official API using release
\texttt{0813}. Qwen-3.6-Flash scores both collections with the same script, formulas,
and rubrics.

\begin{table}[!htbp]
\centering\small
\caption{Decoding settings and input/output budgets.}
\label{tab:execution-settings}
\begin{tabular}{P{.28\textwidth}P{.63\textwidth}}
\toprule\rowcolor{EHRInk!5}Setting & Implementation \\
\midrule
Prompt / output & Chinese task instructions; plain-text JD generation, task-specific JSON otherwise. Retrieval and ranking request all candidate IDs. \\
\addlinespace[3pt]
Temperature (runs) & 0.0; Kimi-K3 omits this parameter and uses the endpoint default. \\
\addlinespace[3pt]
Reasoning (runs) & Thinking disabled where configurable; GLM-5.3 uses low reasoning effort. \\
\addlinespace[3pt]
Output budget (runner) & Default 4,096 tokens outside ranking; ranking uses $\max(16{,}384,\,60N+8{,}192)$ for $N$ candidates. \\
\addlinespace[3pt]
Input truncation & Occupation: 3,500 characters; PJF/PJFA: 3,000 per text; retrieval candidate experience: default 300; ranking candidate profile: 400. \\
\bottomrule
\end{tabular}
\end{table}

The result manifest maps leaderboard names to per-sample export folders for
reproducing the diagnostic panels. It records the judge separately from evaluated models.

\subsection{Generation Rubrics and Component Interpretation}
\label{app:judge}

The two generation tasks combine complementary measurements. Character-level ROUGE-L
compares wording with a reference. PJFA additionally evaluates the structured verdict
against its task label. Qwen-3.6-Flash scores generated content using the operational
rubrics in Table~\ref{tab:rubric-dimensions}. These components retain separate values
in the result exports and in Table~\ref{tab:full-generation}.

\begin{table}[!htbp]
\centering\fontsize{8.5}{10.5}\selectfont\setlength{\tabcolsep}{4pt}
\caption{Rubric dimensions and their scoring scales in the rubric judge.}
\label{tab:rubric-dimensions}
\begin{tabular}{@{}P{.11\textwidth}P{.52\textwidth}P{.32\textwidth}@{}}
\toprule\rowcolor{EHRInk!5}Task & Dimensions & Score formation \\\midrule
JDG & Structure; alignment with job requirements; professionalism; actionable requirements; concision; consistency; attractiveness & Each dimension 0--3; weights 3, 3, 2, 2, 1, 2, 2; normalized to 0--100. \\
\addlinespace[3pt]
PJFA & Reasonableness of strengths/weaknesses; supporting evidence; clarity of structure; professionalism & Four dimensions of 0--25; total 0--100. \\
\bottomrule\end{tabular}
\end{table}

For PJFA, the normal parsed path passes \texttt{analysis\_text} to the rubric judge;
the structured \texttt{verdict} is scored separately. Verdict credit requires both
successful extraction under the shared parsing protocol and agreement with the
reference. Failed extraction or a nonmatching verdict gives $V=0$; even maximal
text components then yield at most 50 composite points. A raw-text fallback applies
when the analysis field is empty or unavailable. The explanation itself can contain a
conclusion, but text credit cannot restore the lost verdict component or establish
JSON compliance. Accordingly, reference-verdict agreement and rubric quality characterize
different aspects of an output; their conjunction is visible in the component and
conditional-score tables (Appendix~\ref{app:diagnostics}).

The judge receives the generated text (up to 2,500 characters). PJFA also includes
up to 2,500 reference-text characters and source context assembled from the first
800 characters of each resume and job, then truncated jointly to 1,200 characters.
JDG receives up to 1,200 characters of its structured job constraints. These are
judge-side budgets: source context is shortened for 400/872 PJFA records (45.9\%).
The reference explanation is visible to the judge, so its assessment is
reference-conditioned. The evaluated-system input budgets are given in
Table~\ref{tab:execution-settings}. The supplied scorer fixes these interfaces across
systems, while retaining verdict accuracy as a separate decision measure.

\section{Full External Results and Scored-Set Coverage}
\label{app:full-results}
\label{app:coverage}

\begin{table}[htbp]
\centering\fontsize{8}{10}\selectfont\setlength{\tabcolsep}{2.5pt}
\caption{Complete external task scores. Bold / underline mark the best / second-best
 distinct displayed values, including ties. Abbreviations follow
 Table~\ref{tab:task-inventory}.}
\begin{tabular}{@{}lrrrrrrrrrrrr@{}}
\toprule
 & \multicolumn{2}{c}{\textcolor{EHRGen}{Gen}} & \textcolor{EHRExt}{Ext} & \multicolumn{3}{c}{\textcolor{EHRCls}{Cls}} & \textcolor{EHRRet}{Ret} & \multicolumn{3}{c}{\textcolor{EHRSTS}{STS}} & \multicolumn{2}{c}{\textcolor{EHRRank}{Rank}} \\
\cmidrule(lr){2-3}\cmidrule(lr){4-4}\cmidrule(lr){5-7}\cmidrule(lr){8-8}\cmidrule(lr){9-11}\cmidrule(lr){12-13}
\rowcolor{EHRInk!5}System & JDG & PJFA & QP & JC & TC & PJF & TR & SimT & SimJ & Dedup & J2T & T2J \\
\midrule\TableOpenTaskRows\bottomrule
\end{tabular}
\end{table}

All twelve systems have the task counts in Table~\ref{tab:task-inventory}, with
two MiniMax-M3 refusals documented below. Ranking validity is assessed separately
in Appendix~\ref{app:diagnostics}.

\paragraph{Refusals and scored-set coverage.}
MiniMax-M3 declines two inputs: \texttt{post\_class::000488} in JC and
\texttt{sim\_post::000067} in SimJ. Its scored counts are therefore 1,063/1,064
and 855/856, respectively. All other systems return outputs for these two inputs,
and every other model--task set is complete. The task scorers omit these two
absent predictions from the main scores.

Table~\ref{tab:fullset} also counts both refusals as failures: zero PathScore for
JC and a true-class error for SimJ, with macro-F1 recomputed from aggregate counts.
The change is less than 0.1 point per affected task and leaves the Overall ordering
unchanged.

\begin{table}[H]
\centering\small\setlength{\tabcolsep}{10pt}
\caption{MiniMax-M3 scores with refusals omitted or counted as failures.}
\label{tab:fullset}
\begin{tabular}{lrrr}\toprule
\rowcolor{EHRInk!5}Task & Scored / total & Main score & Refusals as failures \\\midrule
\ReviewSupportCoverageRows
\bottomrule\end{tabular}
\end{table}

\section{Component and Execution Diagnostics}
\label{app:diagnostics}

\begin{figure}[!htbp]
\centering\includegraphics[width=\textwidth]{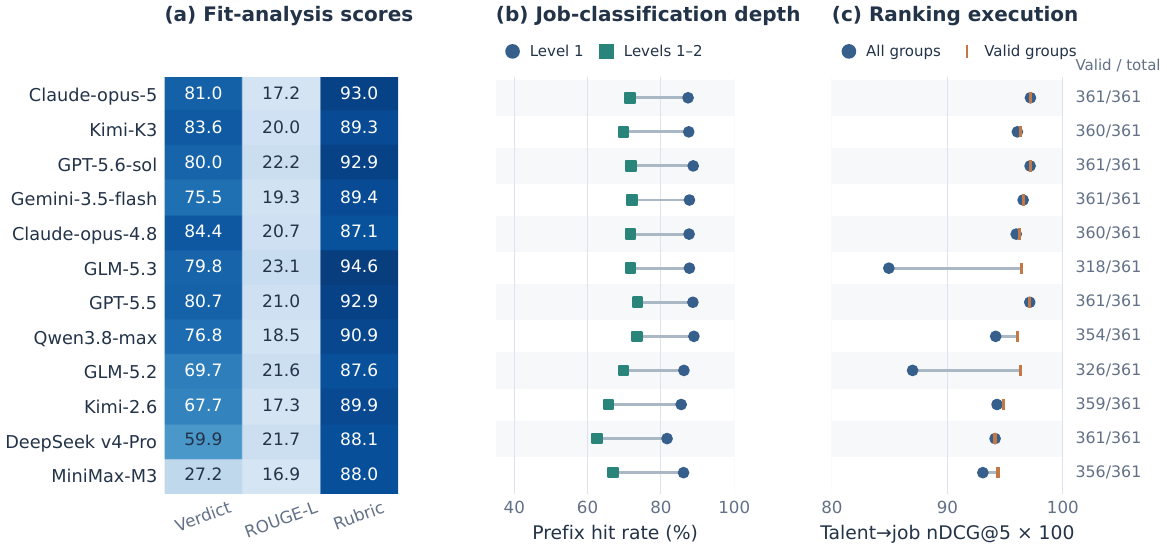}
\caption{Twelve-system diagnostics in Overall order. (a) PJFA components (0--100; $n=872$).
 (b) JC prefix hit rates ($n=1{,}064$; MiniMax-M3: $1{,}063$).
 (c) T2J nDCG@5 on all 361 groups versus valid groups, with valid/total counts.}
\label{fig:diagnostics}
\end{figure}

Figure~\ref{fig:diagnostics} summarizes generation components, occupation depth, and
ranking validity; the following tables give their full task-level breakdowns.

GLM-5.3 has a higher PJFA composite than Claude-opus-5 (65.75 versus 64.25) and
lower verdict accuracy (79.82 versus 80.96). Higher ROUGE-L and rubric scores
account for the difference in composite scores.

For GLM-5.3 relative to Claude-opus-5, the six category differences divided by six
sum to $-0.985$ Overall points from displayed scores: Generation contributes
$+0.378$ and Ranking $-1.057$. This is a decomposition of the fixed aggregation rule.

\begin{table}[htbp]
\centering\fontsize{8}{9.8}\selectfont\setlength{\tabcolsep}{4pt}
\caption{Generation components (0--100). JDG weights ROUGE-L/rubric by 0.6/0.4;
 PJFA weights verdict/ROUGE-L/rubric by 0.5/0.3/0.2.}
\label{tab:full-generation}
\begin{tabular}{lrrrrrrr}\toprule
 & \multicolumn{3}{c}{\textcolor{EHRGen}{\textbf{Job description generation}}} & \multicolumn{4}{c}{\textcolor{EHRGen}{\textbf{Person--job fit analysis}}} \\
\cmidrule(lr){2-4}\cmidrule(lr){5-8}
\rowcolor{EHRInk!5}System & ROUGE-L & Rubric & Score & Verdict & ROUGE-L & Rubric & Score \\
\midrule\TableGenerationComponentRows\bottomrule\end{tabular}
\end{table}

At level $k$, the eligible set contains records with reference depth at least $k$.
All systems correctly match the six depth-three TC records; Table~\ref{tab:occupation-prefix}
reports the first two TC levels and all three JC levels.

\begin{table}[htbp]
\centering\fontsize{8}{10}\selectfont\setlength{\tabcolsep}{3pt}
\caption{Occupation prefix accuracy (percent), with eligible counts in parentheses.
 A hit requires the full prefix through that level.}
\label{tab:occupation-prefix}
\begin{tabular}{@{}lrrrrr@{}}\toprule
 & \multicolumn{3}{c}{Job classification} & \multicolumn{2}{c}{Talent classification} \\
\cmidrule(lr){2-4}\cmidrule(lr){5-6}
\rowcolor{EHRInk!5}System & L1 & L1--2 & L1--3 & L1 & L1--2 \\
\midrule\TableHierarchyRows\bottomrule\end{tabular}
\end{table}

\begin{table}[htbp]
\centering\scriptsize
\caption{Groups satisfying the complete-permutation ranking protocol. Main nDCG@5
 averages over all groups, with zero for groups failing the protocol.}
\begin{tabular}{lrr}\toprule\rowcolor{EHRInk!5}System & Job$\rightarrow$talent & Talent$\rightarrow$job \\
\midrule\TableRankingValidityRows\bottomrule\end{tabular}
\end{table}

GLM-5.3 fails the complete-list requirement on 43 of 361 T2J groups, which
receive zero credit. Its official all-group nDCG@5 is 84.92;
Figure~\ref{fig:diagnostics}c reports valid-group scores.

\FloatBarrier
\begin{samepage}
\subsection{Joint Reading of Verdict and Explanation Scores}
\label{app:conditional-generation}

For each system, Table~\ref{tab:conditional-generation} groups the 872 PJFA outputs
by the scored verdict: reference-matching ($V=100$) and reference-nonmatching
($V=0$). It reports the
rubric mean for both subsets. The rubric evaluates the explanation text under the
criteria in Appendix~\ref{app:judge}; the verdict component measures the structured
decision. The composite assigns half its weight to verdict agreement, ensuring that the
decision contributes directly. Nonmatches include verdicts unavailable under the
required JSON interface and extracted verdicts that disagree with the reference.\par
\end{samepage}

\begin{table}[H]
\centering\fontsize{8.5}{10.5}\selectfont\setlength{\tabcolsep}{6pt}
\caption{PJFA rubric conditional on parsed verdict agreement. Each system's two
 subsets sum to 872 outputs; rubric scores are on a 0--100 scale.}
\label{tab:conditional-generation}
\begin{tabular}{lrrrr}\toprule
 & \multicolumn{2}{c}{Reference-matching} & \multicolumn{2}{c}{Reference-nonmatching} \\
\cmidrule(lr){2-3}\cmidrule(lr){4-5}
\rowcolor{EHRInk!5}System & $n$ & Mean rubric & $n$ & Mean rubric \\\midrule
\ReviewSupportGenerationConditionalRows
\bottomrule\end{tabular}
\end{table}

Across all systems, the mean rubric on reference-nonmatching verdicts is
79.74--92.56. Joint component reporting distinguishes explanation ratings from
successful delivery of a matching verdict.

\subsection{PJFA JSON Output Compliance}
\label{app:pjfa-json}

Table~\ref{tab:pjfa-json} audits all 872 nonempty PJFA responses per system using
the same benchmark JSON parser. It attempts code-fenced JSON, the full response,
and an embedded object from the first opening to the last closing brace, each
requiring successful JSON decoding. The verdict-availability column measures
extraction through this parser. A parsing failure receives zero for the verdict
component; explanation scoring can still use the raw-text fallback.

\begin{table}[H]
\centering\small\setlength{\tabcolsep}{6pt}
\caption{PJFA output parsing under the shared evaluation protocol. Verdict
 availability measures successful extraction through the benchmark JSON parser.}
\label{tab:pjfa-json}
\begin{tabular}{lrrrr}\toprule
\rowcolor{EHRInk!5}System & Nonempty outputs & JSON parsed & Parse rate (\%) & Verdict available (\%) \\
\midrule
Claude-opus-5 & 872 & 858 & 98.39 & 98.39 \\
Kimi-K3 & 872 & 870 & 99.77 & 99.77 \\
GPT-5.6-sol & 872 & 872 & 100.00 & 100.00 \\
Gemini-3.5-flash & 872 & 856 & 98.17 & 98.17 \\
Claude-opus-4.8 & 872 & 869 & 99.66 & 99.66 \\
GLM-5.3 & 872 & 871 & 99.89 & 99.89 \\
GPT-5.5 & 872 & 872 & 100.00 & 100.00 \\
Qwen3.8-max & 872 & 872 & 100.00 & 100.00 \\
GLM-5.2 & 872 & 871 & 99.89 & 99.89 \\
Kimi-2.6 & 872 & 853 & 97.82 & 97.82 \\
DeepSeek v4-Pro & 872 & 872 & 100.00 & 100.00 \\
\rowcolor{EHRInk!5}MiniMax-M3 & 872 & 298 & 34.17 & 34.17 \\
\bottomrule\end{tabular}
\end{table}

MiniMax-M3 yields decodable JSON for 298/872 responses (34.17\%), compared with
97.82--100\% for the other systems. An audit using regular-expression extraction
finds one unique verdict in all 872 responses without requiring valid enclosing
JSON; 611 verdicts match the reference (70.07\%). These findings identify JSON-format
compliance as a major source of verdict-score loss. Table~\ref{tab:conditional-generation}
groups outputs by the shared scoring protocol, under which available explanation
text can still receive rubric credit when verdict extraction fails.

\section{Common-Subset Paired Diagnostics}
\label{app:paired}

GLM-5.3, DeepSeek v4-Pro, and MiniMax-M3 are compared with Claude-opus-5 to examine
generation, binary-relation judgments, and ranking performance.
All twelve tasks are retained for each pair. Differences are comparator minus
Claude-opus-5; scores and intervals use the intersection of scored sample IDs.

Records sharing observed job/person IDs, including pool candidates, form connected
components on the full task before restriction to the common subset. Records without
entity IDs form singletons. We resample represented components with replacement,
preserving record multiplicities and complete query groups. The analysis supplement
records identification coverage and component sizes. Reused candidates connect the
talent-to-job graph into one component; its paired difference is descriptive.

Percentile 95\% intervals use 10,000 replicates with seed 20260916. Each replicate
recomputes F1 from class/span counts, balanced accuracy from class recalls with the
scorer's invalid-output mapping, and other scores from record/group totals.
Single-class balanced-accuracy replicates are omitted and counted. These exploratory intervals
describe the common subset and observed identity links.

\begingroup\fontsize{8.5}{10}\selectfont
\begin{longtable}{@{}llrrrr@{}}
\caption{Paired differences against Claude-opus-5. $n$: common records/groups;
 $C$: observed-entity components. A dash denotes a single component.}\\
\toprule\rowcolor{EHRInk!5}Comparator & Task & $n$ & $C$ & Difference & 95\% interval \\\midrule
\endfirsthead
\multicolumn{6}{l}{\small Table~\thetable: Paired differences (continued)}\\
\toprule\rowcolor{EHRInk!5}Comparator & Task & $n$ & $C$ & Difference & 95\% interval \\\midrule
\endhead
\TablePairedRows
\bottomrule
\end{longtable}
\endgroup

\subsection{Comparisons among the Leading Overall Systems}
\label{app:top-paired}

Table~\ref{tab:top-paired} compares Claude-opus-5 with the next three systems in
Overall order, retaining every task for each pair. Differences are Claude minus the
named comparator. These task-specific comparisons use the same component-resampling
procedure as above, with 10,000 replicates and seed 20260921. Intervals are exploratory,
pointwise 95\% percentile intervals; their interpretation concerns each task and its
observed entity links. The complete collection of comparisons is reported together.

\begingroup\fontsize{8}{9.5}\selectfont
\begin{longtable}{@{}llrrrr@{}}
\caption{Leading-system paired task differences. $n$: common records/groups;
 $C$: observed-entity components. A dash denotes one connected component.}
\label{tab:top-paired}\\
\toprule\rowcolor{EHRInk!5}Comparator & Task & $n$ & $C$ & Difference & 95\% interval \\\midrule
\endfirsthead
\multicolumn{6}{l}{\small Leading-system paired task differences (continued)}\\
\toprule\rowcolor{EHRInk!5}Comparator & Task & $n$ & $C$ & Difference & 95\% interval \\\midrule
\endhead
\ReviewSupportTopPairedRows
\bottomrule
\end{longtable}
\endgroup

Overall retains the six-family equal-weight definition. A separate leave-one-family-out
view in the analysis supplement illustrates the contribution of capability coverage:
Claude-opus-5 ranks first for four of the six five-family means; omitting Retrieval puts
Kimi-K3 first, while omitting Ranking puts GLM-5.3 first. These operational emphases
support reporting the full capability profile.

\begin{table}[htbp]\centering\small
\caption{Leading five-family mean after omitting each capability family.}
\begin{tabular}{lrr}\toprule
Omitted family & Leading system & Five-family mean \\\midrule
Generation & Claude-opus-5 & 78.428 \\
Extraction & Claude-opus-5 & 71.030 \\
Classification & Claude-opus-5 & 75.194 \\
Retrieval & Kimi-K3 & 76.770 \\
STS & Claude-opus-5 & 73.414 \\
Ranking & GLM-5.3 & 72.430 \\
\bottomrule\end{tabular}\end{table}

\begin{samepage}
\section{Source and Candidate Conditions}
\label{app:slices}

The analysis retains within-task source, occupation, reference-depth, query-style,
construction-type, and candidate-condition slices wherever those metadata occur.
Missing metadata are recorded through per-field coverage rather than inferred.
Candidate-size bins are 1--8, 9--16, and 17 or more; positive-count bins are one versus
multiple; the positive-rate split is 0.125, and available same-second-level occupation
ratios are split at 0.5. These fixed descriptive bins are applied to the benchmark data.
Both labels must occur for reporting slice balanced accuracy or binary macro-F1; single-class
slices report positive recall or negative rejection instead. All denominators are
retained. Complete results, including all occupation strata, are in the machine-readable
analysis supplement; Tables~\ref{tab:source-slices}--\ref{tab:source-slices-4} give
every multi-source task's source breakdown for all twelve systems.\par
\end{samepage}

The candidate census finds 104 retrieval pools of size 50, 47 of size 12, and 33 of
size eight. The two ranking directions have one positive per group, whereas retrieval
has between one and 28. Job-to-talent positive input positions 1--10 have counts
13, 16, 20, 10, 22, 13, 7, 4, 6, and 5; talent-to-job positions 1--8 have counts
52, 48, 37, 52, 53, 35, 34, and 50. These counts determine the input-order reference scores in
Table~\ref{tab:baseline-source}. The strict permutation and top-ten protocols
likewise determine how each candidate ordering is scored.

Source panels in Tables~\ref{tab:source-slices}--\ref{tab:source-slices-4}
retain every system and denominator. ``Public'' denotes original public-dataset
records, ``Public JD'' public job descriptions, ``Derived'' public-data-derived,
and ``Synthetic'' controlled synthetic records.

Talent classification makes the associated input conditions concrete: the 91 public
records have median input length 219 characters and depth-two labels throughout;
the 70 synthetic records have median length 979, with 64 depth-two and six depth-three
labels. Across twelve systems, their score ranges are 38.46--44.51 and 85.00--99.29,
respectively. These source profiles reflect jointly varying content and label
conditions; the tables report every system's scores.

\begin{table}[!htbp]
\centering\fontsize{8}{10}\selectfont\setlength{\tabcolsep}{3.5pt}
\caption{Source scores for JDG, PJFA, TC, TR. Header counts give stratum sizes.}
\label{tab:source-slices}
\label{tab:source-slices-2}
\begin{tabular}{lrrrrrrrr}\toprule
 & \multicolumn{2}{c}{\textbf{JDG}} & \multicolumn{2}{c}{\textbf{PJFA}} & \multicolumn{2}{c}{\textbf{TC}} & \multicolumn{2}{c}{\textbf{TR}} \\
\rowcolor{EHRInk!5}System & \shortstack{Public\\(500)} & \shortstack{Public JD\\(441)} & \shortstack{Derived\\(500)} & \shortstack{Synthetic\\(372)} & \shortstack{Public\\(91)} & \shortstack{Synthetic\\(70)} & \shortstack{Public\\(104)} & \shortstack{Derived\\(80)} \\\midrule
Claude-opus-5 & 43.46 & 48.48 & 57.41 & 73.44 & 40.66 & 97.86 & 45.55 & 86.53 \\
Kimi-K3 & 42.65 & 46.87 & 58.95 & 74.72 & 41.21 & 97.14 & 42.86 & 86.53 \\
GPT-5.6-sol & 41.14 & 46.89 & 57.18 & 76.17 & 40.66 & 99.29 & 43.08 & 83.92 \\
Gemini-3.5-flash & 42.05 & 47.54 & 53.61 & 71.85 & 39.56 & 99.29 & 41.15 & 85.74 \\
Claude-opus-4.8 & 41.89 & 46.45 & 58.93 & 75.09 & 42.86 & 97.14 & 40.09 & 86.91 \\
GLM-5.3 & 47.60 & 50.29 & 57.34 & 77.06 & 41.76 & 94.29 & 40.61 & 85.10 \\
GPT-5.5 & 40.52 & 46.27 & 57.56 & 75.63 & 39.01 & 98.57 & 42.70 & 84.69 \\
Qwen3.8-max & 43.94 & 48.60 & 53.28 & 74.05 & 40.11 & 97.14 & 42.98 & 85.95 \\
GLM-5.2 & 43.73 & 48.24 & 46.68 & 75.22 & 43.96 & 94.29 & 42.02 & 85.86 \\
Kimi-2.6 & 42.34 & 47.00 & 46.01 & 71.75 & 44.51 & 91.43 & 41.20 & 85.53 \\
DeepSeek v4-Pro & 43.10 & 48.46 & 38.28 & 75.31 & 40.11 & 85.71 & 40.47 & 84.17 \\
MiniMax-M3 & 41.92 & 46.03 & 27.91 & 47.49 & 38.46 & 85.00 & 41.10 & 85.41 \\
\bottomrule\end{tabular}
\end{table}

\begin{table}[!htbp]
\centering\fontsize{8}{10}\selectfont\setlength{\tabcolsep}{3.5pt}
\caption{Source scores for SimT, SimJ, Dedup, J2T. Header counts give stratum sizes.}
\label{tab:source-slices-3}
\label{tab:source-slices-4}
\begin{tabular}{lrrrrrrrr}\toprule
 & \multicolumn{2}{c}{\textbf{SimT}} & \multicolumn{2}{c}{\textbf{SimJ}} & \multicolumn{2}{c}{\textbf{Dedup}} & \multicolumn{2}{c}{\textbf{J2T}} \\
\rowcolor{EHRInk!5}System & \shortstack{Derived\\(500)} & \shortstack{Synthetic\\(311)} & \shortstack{Derived\\(540)} & \shortstack{Synthetic\\(316)} & \shortstack{Derived\\(616)} & \shortstack{Synthetic\\(254)} & \shortstack{Derived\\(110)} & \shortstack{Synthetic\\(6)} \\\midrule
Claude-opus-5 & 65.85 & 53.89 & 74.09 & 94.64 & 99.51 & 95.67 & 73.65 & 74.87 \\
Kimi-K3 & 58.33 & 72.78 & 71.92 & 98.15 & 97.89 & 89.32 & 72.42 & 79.36 \\
GPT-5.6-sol & 64.14 & 56.95 & 72.77 & 99.07 & 99.03 & 93.69 & 73.05 & 81.55 \\
Gemini-3.5-flash & 67.90 & 72.81 & 74.95 & 91.97 & 97.08 & 94.88 & 74.16 & 76.03 \\
Claude-opus-4.8 & 59.44 & 65.49 & 73.75 & 96.67 & 98.05 & 85.29 & 73.59 & 74.87 \\
GLM-5.3 & 65.81 & 70.05 & 72.53 & 96.77 & 97.73 & 92.52 & 73.37 & 72.06 \\
GPT-5.5 & 63.72 & 55.25 & 72.34 & 98.13 & 98.54 & 90.50 & 70.93 & 79.36 \\
Qwen3.8-max & 53.74 & 57.11 & 74.41 & 89.71 & 95.77 & 81.63 & 72.78 & 69.24 \\
GLM-5.2 & 67.88 & 69.57 & 75.03 & 94.12 & 96.92 & 92.89 & 70.34 & 58.73 \\
Kimi-2.6 & 67.27 & 70.28 & 72.74 & 95.55 & 96.43 & 88.06 & 70.14 & 68.72 \\
DeepSeek v4-Pro & 69.09 & 75.30 & 73.98 & 90.29 & 96.26 & 72.32 & 67.00 & 73.21 \\
MiniMax-M3 & 62.76 & 69.57 & 71.86 & 93.81 & 96.75 & 67.56 & 68.53 & 74.87 \\
\bottomrule\end{tabular}
\end{table}

\subsection{PJF Construction Strata and a Supervised Lexical Probe}
\label{app:pjf-construction}

PJF contains 900 behavior-derived pairs (478 satisfaction positives, 207
browsing-without-application negatives, and 215 application-without-satisfaction
negatives) and 486 reviewed constructed pairs (162 positives and 324 negatives).
Table~\ref{tab:pjf-construction} reports balanced accuracy from final labels and
saved predictions within each stratum. Full-task balanced accuracy pools the class
counts across strata before averaging positive and negative recall.

\begin{table}[htbp]\centering\fontsize{8}{10}\selectfont
\caption{PJF performance by construction stratum (percent). Behavior TPR/TNR
 use 478 positive/422 negative final labels.}
\label{tab:pjf-construction}
\begin{tabular}{lrrrr}\toprule
System & Behavior BA & Constructed BA & Behavior TPR & Behavior TNR \\\midrule
\ExpertPjfConstructionRows
\bottomrule\end{tabular}\end{table}

The probe combines character 1--2-gram TF-IDF (sublinear term frequency, minimum
document frequency 2) with logistic regression ($C=1$, balanced class weights,
at most 3,000 iterations).

\begin{samepage}
Vocabulary and IDF are fitted within each training fold.
Five-fold cross-validation groups by resume ID or job
ID, with three deterministic group relabelings (seeds 0, 1, 2); each feature
configuration uses the same folds within a grouping. Scores pool out-of-fold
predictions per seed and then average seeds. A fixed probability threshold of
0.5 gives balanced accuracy. Seed standard deviations describe split variation,
not confidence intervals. No metadata, IDs, or construction tags are features.\par
\end{samepage}

\begin{table}[htbp]\centering\small
\caption{Supervised PJF probe ($n=900$): AUC from held-out probabilities; BA at a 0.5 threshold.}
\label{tab:pjf-cv}
\begin{tabular}{llrrr}\toprule
Held-out entity & Text features & AUC & AUC SD & BA (\%) \\\midrule
\ExpertPjfCvRows
\bottomrule\end{tabular}\end{table}

There are 764 unique resumes and 886 jobs; median text lengths are 154.5 and
363 characters. Job-only features exceed both-text features under either grouping,
indicating predictive associations in job text. Resume-only features provide
a complementary control. Each split holds out its named entity type; the probe
characterizes associations within the benchmark's behavioral collection.

A separate unsupervised character-bigram cosine check fits IDF on all 1,800
unlabeled texts. Its AUC is 0.400 overall, 0.469 against browsing negatives, and
0.334 against application-without-satisfaction negatives. The supervised and
unsupervised controls together distinguish learnable behavioral associations from
surface overlap. The two PJF strata therefore support evaluation against both
interaction evidence and reviewed constructed judgments under a common task interface.

\subsection{Relation Coverage, Selectivity, and Construction Difficulty}
\label{app:operating-points}

\begin{figure}[htbp]\centering
\includegraphics[width=\textwidth]{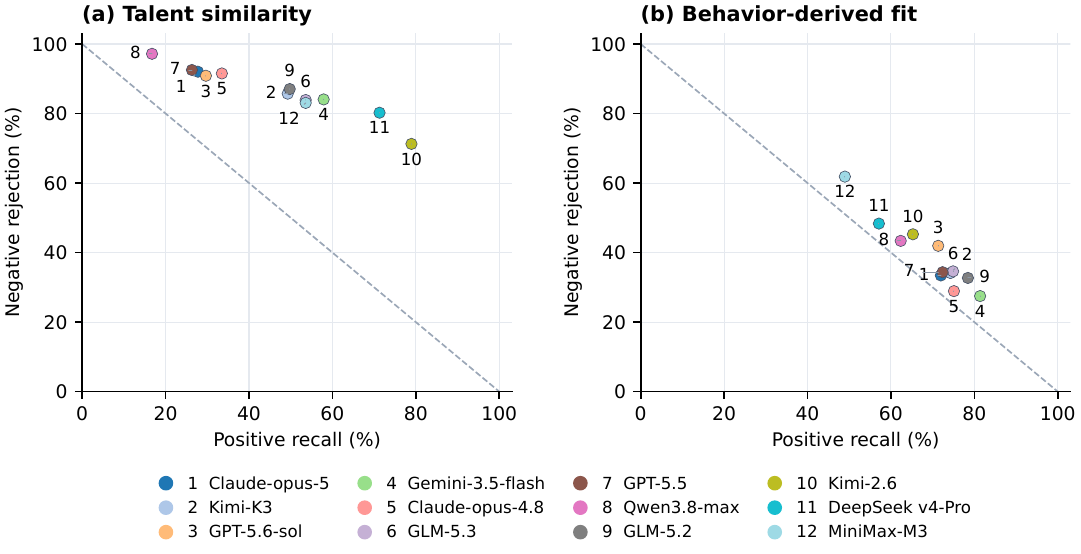}
\caption{Positive recall and negative rejection for (a) SimT ($n=811$) and
 (b) behavior-derived PJF ($n=900$). Dashed lines mark 50\% balanced accuracy;
 numbers identify systems.}
\label{fig:operating-points}
\end{figure}

SimT evaluates related-profile discovery using shared-job and occupation-based
construction followed by label review. Positive recall ranges from 16.75\% to
78.95\%, while negative rejection ranges from 71.26\% to 97.18\%.
Figure~\ref{fig:operating-points} shows that systems with greater positive recall
also tend to accept more negative pairs. On SimJ, rejection of the 141 same-category
pairs with final negative labels ranges
from 4.96\% to 22.70\%, versus 99.50--100\% on the 199 cross-category negatives.
Thus the narrow SimJ range masks large differences in rejection between the two
negative strata.

HR-reviewed labels determine scoring; construction tags record sampling intent.
SimT's 167 shared-job positive seeds yield 105 positive and 62 negative final labels,
which also define the class-conditional diagnostics (Table~\ref{tab:difficulty-strata}).

\begin{table}[htbp]\centering\fontsize{8}{10}\selectfont
\caption{Score ranges across twelve systems by construction condition.
 Binary rows use final-label recall/rejection; QP uses span F1.}
\label{tab:difficulty-strata}
\begin{tabular}{llrr}\toprule
Task & Stratum and statistic & Total $n$ & System range \\\midrule
\ExpertDifficultyStrataRows
\bottomrule\end{tabular}\end{table}

\subsection{Construction-Model and Judge-Family Overlap}
\label{app:generator-overlap}

Table~\ref{tab:generator-overlap} compares the two resume generators with other
systems across all five multi-source tasks containing controlled-synthetic records.
Claude-opus-5 and GPT-5.6-sol rank twelfth and tenth on synthetic SimT, while GPT
leads synthetic SimJ and ties for first on TC, and Claude leads Dedup. The generators
therefore show task-dependent strengths, without a consistent advantage across
synthetic strata. These are source-level comparisons; the metadata do not attribute
each record to a generator.

\begin{table}[htbp]\centering\fontsize{8}{10}\selectfont
\caption{Controlled-synthetic scores of the two resume generators versus the other
 ten systems. Parentheses give ranks among all twelve systems, with ties.}
\label{tab:generator-overlap}
\begin{tabular}{lrrrr}\toprule
Task & Claude-opus-5 & GPT-5.6-sol & Others' mean & Others' range \\\midrule
\ExpertGeneratorOverlapRows
\bottomrule\end{tabular}\end{table}

The rubric judge is Qwen-3.6-Flash. The evaluated Qwen3.8-max has JDG/PJFA rubric
means of 85.75/90.91, below GLM-5.3's 91.12/94.64. The evaluated judge-family model
leads neither rubric.

\subsection{Candidate-Task Baselines and Paired Gains}
\label{app:baselines}

Three offline reference methods characterize candidate conditions and contextualize
LLM scores using the fixed task labels and pools.
\emph{Input order} returns the supplied list unchanged. \emph{Uniform permutation}
reports the exact expected score over all candidate permutations, conditional on each
pool's size and relevant-item count. \emph{Character-bigram TF-IDF} ranks candidates
by cosine similarity to the anchor, with lowercasing, collapsed whitespace, raw term
frequency, smoothed IDF $\log((1+N)/(1+\mathrm{df}))+1$, and L2 normalization.
IDF is fitted to the unique prompt-visible texts of each task, an unlabeled transductive
setting; neither labels, reference answers, identifiers, nor source tags are features.
Similarity ties retain input order. The method uses fixed settings, without score-based
tuning, and always returns a complete permutation.

Text fields follow the evaluation runner: ranking anchors use up to 1,500 characters and
candidates 400; retrieval uses job-title/category/requirement fields, a job description
of up to 1,500 characters, and candidate experience of up to 300 characters. Labels
enter only the final metric computation. TR averages Recall@3, Recall@10, and MRR@10;
both ranking tasks use nDCG@5. Single-positive ranking measures placement of the
relevant candidate, while the complete-permutation requirement measures list execution.

\begin{table}[!htbp]
\centering\fontsize{8}{10}\selectfont\setlength{\tabcolsep}{5pt}
\caption{Baseline and model scores within each source. The model is the overall task
 leader: Claude-opus-5 for TR/T2J and Gemini-3.5-flash for J2T. $\Delta$: model minus TF-IDF.}
\label{tab:baseline-source}
\begin{tabular}{llrrrrrr}\toprule
\rowcolor{EHRInk!5}Task & Source & $n$ & Uniform & Input & TF-IDF & Model & $\Delta$ \\\midrule
\ReviewSupportBaselineSourceRows
\bottomrule\end{tabular}
\end{table}

\begin{table}[!htbp]
\centering\fontsize{8}{10}\selectfont\setlength{\tabcolsep}{4pt}
\caption{Paired gains over TF-IDF on all groups. W/T/L counts higher/equal/lower
 group scores at four decimals (0--1 scale). Intervals use observed-entity
 components; T2J has one component.}
\label{tab:baseline-paired}
\begin{tabular}{llrrr}\toprule
\rowcolor{EHRInk!5}Task & System & W/T/L & $\Delta$ & 95\% interval \\\midrule
\ReviewSupportBaselinePairedRows
\bottomrule\end{tabular}
\end{table}

\begin{samepage}
The gains use common complete group sets and the component-resampling procedure from
Appendix~\ref{app:paired}, with 10,000 replicates and seed 20260921. The model for each
task is selected by its reported task score; these are exploratory comparisons of the
observed task leaders. Both sources in TR and J2T show positive mean model gains.
T2J's 75 improved, 276 tied, and ten lower-scoring groups show how model gains coexist
with strong lexical performance. The six synthetic J2T groups are reported with their
own denominator. All methods use the same supplied pools; source comparisons describe
these input and label conditions.\par
\end{samepage}

\begin{figure}[!ht]
\centering\includegraphics[width=\textwidth]{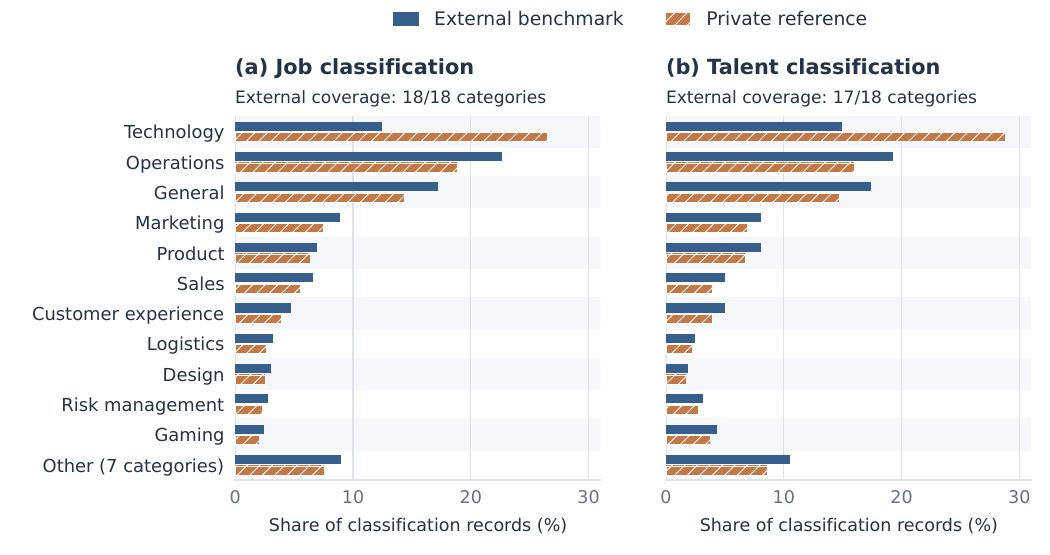}
\caption{Occupation shares across eighteen categories. Categories below 3\% in all
 four cohorts form Other. Tables~\ref{tab:population} and~\ref{tab:occupation-counts}
 give cohort sizes and counts.}
\label{fig:occupation-coverage}
\end{figure}

\begin{samepage}
\section{Operational Results and Correspondence Details}
\label{app:comparison}
\subsection{Population Statistics and Measurement Units}
\label{app:population-definitions}

Figure~\ref{fig:occupation-coverage} compares the two classification cohorts using
the shared eighteen-category first level. The 1,064 external job records cover all
eighteen categories, and the 161 talent records cover seventeen; the internal cohorts
cover all eighteen. The external proportions are recomputed from benchmark reference
labels, and the internal proportions come from the construction team's aggregate
plotting supplement. Both comparisons yield TV = 0.140 to three decimal places.\par
\end{samepage}

\begin{table}[H]
\centering\fontsize{8}{10}\selectfont\setlength{\tabcolsep}{4pt}\renewcommand{\arraystretch}{1.06}
\caption{External and private-reference populations. TV measures occupation
 distribution distance; ranking pool means are J2T/T2J.}
\label{tab:population}
\begin{tabular}{lrrl}
\toprule
\rowcolor{EHRInk!5}\textbf{Property} & \textbf{External} & \textbf{Internal} & \textbf{Comparison} \\
\midrule
Job-classification records & 1,064 & 10,599 & 18/18 classes; TV = 0.140 \\
Talent-classification records & 161 & 7,716 & 17/18 classes; TV = 0.140 \\
\addlinespace[3pt]
Job-text mean / median chars & 526 / 460 & 528 / 423 & Mean difference 0.4\% \\
PJF city coverage (\%) & 92.1 & 100 & 7.9 pp difference \\
PJF education coverage (\%) & 99.9 & 100 & 0.1 pp difference \\
PJF experience coverage (\%) & 97.9 & 100 & 2.1 pp difference \\
\addlinespace[3pt]
Ranking pool means & 7.3 / 8.0 & 8.1 / 6.0 & Task-specific pools \\
\bottomrule
\end{tabular}
\end{table}

\begin{table}[!htbp]
\centering\small\setlength{\tabcolsep}{4pt}
\caption{Measurement definitions for population comparisons.
 Sample counts are external/internal; ranking units are complete groups.}
\label{tab:population-definitions}
\begin{tabular}{P{.23\textwidth}P{.47\textwidth}P{.21\textwidth}}
\toprule\rowcolor{EHRInk!5}Statistic & Measurement & Sample count \\
\midrule
Occupation proportions & First-level reference category, \texttt{cate\_labels[0]};
 TV is half the summed absolute proportion differences. &
 \shortstack[l]{Job: 1,064/10,599\\Talent: 161/7,716} \\
\addlinespace[3pt]
Job-text length & Character count of \texttt{input.post\_text} in job classification. & 1,064/10,599 \\
\addlinespace[3pt]
PJF field coverage & Same regex/rule detector applied to resume input text
 (external: \texttt{input.resume\_info}) for city,
 education, and work-experience content. & 1,386/880 \\
\addlinespace[3pt]
Ranking pool size & Number of entries in \texttt{input.candidates}, summarized per group. &
 \shortstack[l]{J2T: 116/123\\T2J: 361/82} \\
\bottomrule
\end{tabular}
\end{table}

The external job-length mean and median are 525.56 and 460 characters, respectively;
Table~\ref{tab:population} reports rounded values. PJF field coverage measures detected
information in resume text, and candidate groups remain the unit for pool comparisons.

\begin{table}[!htbp]
\centering\fontsize{8.5}{10.5}\selectfont
\setlength{\tabcolsep}{4pt}\renewcommand{\arraystretch}{1.06}
\caption{Complete first-level occupation counts and shares. Counts are external
 records; percentages use each cohort's denominator. $\dagger$ marks the categories
 grouped as Other in Figure~\ref{fig:occupation-coverage}: each is below 3\% in all
 four cohorts.}
\label{tab:occupation-counts}
\begin{tabular}{@{}lrrrrrr@{}}
\toprule
& \multicolumn{3}{c}{\textbf{Job classification}} & \multicolumn{3}{c}{\textbf{Talent classification}} \\
\cmidrule(lr){2-4}\cmidrule(lr){5-7}
\rowcolor{EHRInk!5}\textbf{Occupation} & \textbf{Ext. $n$} & \textbf{Ext. (\%)} & \textbf{Int. (\%)} & \textbf{Ext. $n$} & \textbf{Ext. (\%)} & \textbf{Int. (\%)} \\
\midrule
Technology & 133 & 12.50 & 26.50 & 24 & 14.91 & 28.82 \\
Operations & 241 & 22.65 & 18.87 & 31 & 19.25 & 15.98 \\
General & 183 & 17.20 & 14.33 & 28 & 17.39 & 14.67 \\
Marketing & 95 & 8.93 & 7.48 & 13 & 8.07 & 6.88 \\
Product & 74 & 6.95 & 6.36 & 13 & 8.07 & 6.74 \\
Sales & 70 & 6.58 & 5.49 & 8 & 4.97 & 3.90 \\
Customer experience & 50 & 4.70 & 3.93 & 8 & 4.97 & 3.93 \\
Logistics & 34 & 3.20 & 2.63 & 4 & 2.48 & 2.23 \\
Design & 32 & 3.01 & 2.52 & 3 & 1.86 & 1.71 \\
Risk management & 30 & 2.82 & 2.33 & 5 & 3.11 & 2.77 \\
Gaming & 26 & 2.44 & 2.01 & 7 & 4.35 & 3.77 \\
Data$^{\dagger}$ & 25 & 2.35 & 1.93 & 4 & 2.48 & 1.92 \\
Supply chain$^{\dagger}$ & 23 & 2.16 & 1.83 & 3 & 1.86 & 1.65 \\
Finance$^{\dagger}$ & 14 & 1.32 & 1.11 & 3 & 1.86 & 1.37 \\
Project management$^{\dagger}$ & 13 & 1.22 & 1.05 & 4 & 2.48 & 1.87 \\
Content$^{\dagger}$ & 12 & 1.13 & 0.91 & 2 & 1.24 & 1.21 \\
Advertising$^{\dagger}$ & 5 & 0.47 & 0.43 & 1 & 0.62 & 0.51 \\
Map data$^{\dagger}$ & 4 & 0.38 & 0.29 & 0 & 0.00 & 0.08 \\
\bottomrule
\end{tabular}
\end{table}

\subsection{Scores and Correspondence Statistics}

\begin{table}[!htbp]
\centering\scriptsize
\caption{Private-reference capability scores; bold marks column maxima. The
 task-specific internal suite is excluded from shared-model correspondence.}
\begin{tabular}{lrrrrrrr}\toprule\rowcolor{EHRInk!5}System & \textbf{Overall} & \caphead{EHRGen}{Gen} & \caphead{EHRExt}{Ext} & \caphead{EHRCls}{Cls} & \caphead{EHRRet}{Ret} & \caphead{EHRSTS}{STS} & \caphead{EHRRank}{Rank} \\
\midrule\TablePrivateCategoryRows\bottomrule\end{tabular}
\end{table}

\begin{table}[htbp]
\centering\fontsize{8}{10}\selectfont\setlength{\tabcolsep}{2.5pt}
\caption{Task scores on the private operational reference.}
\begin{tabular}{@{}lrrrrrrrrrrrr@{}}
\toprule
 & \multicolumn{2}{c}{\textcolor{EHRGen}{Gen}} & \textcolor{EHRExt}{Ext} & \multicolumn{3}{c}{\textcolor{EHRCls}{Cls}} & \textcolor{EHRRet}{Ret} & \multicolumn{3}{c}{\textcolor{EHRSTS}{STS}} & \multicolumn{2}{c}{\textcolor{EHRRank}{Rank}} \\
\cmidrule(lr){2-3}\cmidrule(lr){4-4}\cmidrule(lr){5-7}\cmidrule(lr){8-8}\cmidrule(lr){9-11}\cmidrule(lr){12-13}
\rowcolor{EHRInk!5}System & JDG & PJFA & QP & JC & TC & PJF & TR & SimT & SimJ & Dedup & J2T & T2J \\
\midrule\TablePrivateTaskRows\bottomrule\end{tabular}
\end{table}

Spearman correlations use average ranks within the five shared systems on each side.
Direction agreement counts a pair only when both displayed score differences have the
same nonzero sign. All ten pairs remain in each view's denominator, including a tie
on either side. Table~\ref{tab:comparison-full} retains all capability views and
Table~\ref{tab:comparison-tasks} all tasks.

For Overall, under an exploratory exchangeable-order permutation null, exact
positive concordance has one-sided $p=1/120=0.0083$ (two-sided $p=0.0167$).
This describes the ordering under that null, not uncertainty from sampling records
or repeated model calls.

\begin{table}[htbp]\centering\small
\caption{Overall and category correspondence for the five shared systems.}
\label{tab:comparison-full}
\begin{tabular}{lrrrr}\toprule\rowcolor{EHRInk!5}View & Spearman $\rho$ & Same / all & Opposite & Tied \\
\midrule\TableCalibrationRows\bottomrule\end{tabular}
\end{table}
\begin{table}[htbp]\centering\small
\caption{Task correspondence. Each task contributes ten comparisons.}
\label{tab:comparison-tasks}
\begin{tabular}{lrrrr}\toprule\rowcolor{EHRInk!5}Task & Spearman $\rho$ & Same & Opposite & Tied \\
\midrule\TableTaskComparisonRows\bottomrule\end{tabular}
\end{table}

The pooled task comparison has 78 same-direction, 40 opposite-direction, and two tied
pairs at displayed precision, giving 78/120 = 65.0\% with ties retained in the denominator.
These pairs are dependent, and pooling weights tasks equally, whereas Overall weights
capabilities equally.
The displayed ties are Qwen3.8-max/Kimi-2.6
on talent classification and Kimi-K3/Kimi-2.6 on job similarity.
For standardized task scores, each task is centered and scaled within the five models
separately on each side; pooling the twelve standardized pairs of vectors gives
$r=0.494$. Overall pairwise score differences give $r=0.713$. These are descriptive
correspondence statistics. Their interpretation concerns patterns and ordering rather
than equality of numerical score gaps. The internal Overall difference between
Qwen3.8-max and GLM-5.2 is 0.04 points.

\subsection{Public--Private Agreement under Capability Omission}
\label{app:operational_sensitivity}

We assess how the five shared models' public--private ordering changes when one
capability family is removed from both collections. For each of the six omission
conditions, we average the other five reported capability scores with equal
weights and compute Spearman correlation between the two model rankings.
Both sides use the two-decimal capability scores in
\texttt{data/text\_results.json}; we do not substitute higher-precision public
recomputations or infer scores from the reported Overall. Scores are converted to
integer hundredths before summation, and equal sums receive average ranks.

Agreement is $\rho=1.00$ after omitting generation, retrieval, or ranking, and
$\rho=0.90$ after omitting extraction, classification, or STS
(Table~\ref{tab:operational_sensitivity}). Table~\ref{tab:operational_sensitivity_models}
gives every model's mean and rank in all six conditions. Together, these results
show closely aligned aggregate orderings across all six omissions. This descriptive
analysis assesses aggregation sensitivity for the five shared models at the reported
precision; task validity and sampling uncertainty require separate evidence. It
compares both collections, complementing the capability-omission analysis of the
public twelve-model leaderboard. The official Overall scores and rankings are unchanged.
The calculation and its input hash are recorded in
\texttt{scripts/operational\_sensitivity.py} and
\texttt{data/operational\_sensitivity.json}.

\begin{table}[htbp]
\centering\small
\caption{Public--private ranking agreement among five shared models after
omitting the same capability family from both collections. Each remaining
family has equal weight.}
\label{tab:operational_sensitivity}
\begin{tabular}{lr}
\toprule
Omitted capability & Spearman $\rho$ \\
\midrule
Generation & 1.00 \\
Extraction & 0.90 \\
Classification & 0.90 \\
Retrieval & 1.00 \\
STS & 0.90 \\
Ranking & 1.00 \\
\bottomrule
\end{tabular}
\end{table}

\begingroup\small
\begin{longtable}{llrrrr}
\caption{Five-capability means and model ranks underlying
Table~\ref{tab:operational_sensitivity}. Means average two-decimal reported
inputs and are displayed to three decimals to preserve differences introduced
by averaging. Rank 1 is highest; equal sums receive average ranks.}
\label{tab:operational_sensitivity_models}\\
\toprule
 & & \multicolumn{2}{c}{Public} & \multicolumn{2}{c}{Private} \\
\cmidrule(lr){3-4}\cmidrule(lr){5-6}
Omitted & Model & Mean & Rank & Mean & Rank \\
\midrule
\endfirsthead
\multicolumn{6}{c}{Table \thetable{} continued}\\
\toprule
Omitted & Model & Public mean & Rank & Private mean & Rank \\
\midrule
\endhead
\midrule
\multicolumn{6}{r}{Continued on next page}\\
\endfoot
\bottomrule
\endlastfoot
Generation & Kimi-K3 & 78.106 & 1 & 74.312 & 1 \\*
 & Qwen3.8-max & 76.632 & 2 & 73.904 & 2 \\*
 & GLM-5.2 & 76.524 & 3 & 73.670 & 3 \\*
 & Kimi-2.6 & 76.424 & 4 & 73.420 & 4 \\*
 & DeepSeek v4-Pro & 75.154 & 5 & 72.712 & 5 \\
\midrule
Extraction & Kimi-K3 & 70.806 & 1 & 66.078 & 1 \\*
 & Qwen3.8-max & 69.302 & 2 & 65.624 & 3 \\*
 & GLM-5.2 & 69.096 & 3 & 65.630 & 2 \\*
 & Kimi-2.6 & 68.990 & 4 & 65.512 & 4 \\*
 & DeepSeek v4-Pro & 67.350 & 5 & 63.874 & 5 \\
\midrule
Classification & Kimi-K3 & 74.892 & 1 & 70.570 & 1 \\*
 & Qwen3.8-max & 73.192 & 2 & 70.112 & 3 \\*
 & GLM-5.2 & 72.682 & 3 & 70.466 & 2 \\*
 & Kimi-2.6 & 72.586 & 4 & 69.654 & 4 \\*
 & DeepSeek v4-Pro & 71.950 & 5 & 68.760 & 5 \\
\midrule
Retrieval & Kimi-K3 & 76.770 & 1 & 69.316 & 1 \\*
 & Qwen3.8-max & 75.128 & 2 & 68.742 & 2 \\*
 & GLM-5.2 & 74.778 & 3 & 68.614 & 3 \\*
 & Kimi-2.6 & 74.482 & 4 & 68.466 & 4 \\*
 & DeepSeek v4-Pro & 73.228 & 5 & 67.686 & 5 \\
\midrule
STS & Kimi-K3 & 72.862 & 1 & 71.068 & 2 \\*
 & Qwen3.8-max & 72.264 & 2 & 71.316 & 1 \\*
 & GLM-5.2 & 70.568 & 3 & 70.934 & 3 \\*
 & Kimi-2.6 & 70.258 & 4 & 70.426 & 4 \\*
 & DeepSeek v4-Pro & 68.950 & 5 & 68.794 & 5 \\
\midrule
Ranking & Kimi-K3 & 72.254 & 1 & 71.556 & 1 \\*
 & Qwen3.8-max & 70.782 & 3 & 71.372 & 3 \\*
 & GLM-5.2 & 71.322 & 2 & 71.546 & 2 \\*
 & Kimi-2.6 & 70.140 & 4 & 70.622 & 4 \\*
 & DeepSeek v4-Pro & 68.978 & 5 & 69.484 & 5 \\
\end{longtable}
\endgroup

\FloatBarrier
\section{Positioning Against Recruitment Resources}
\label{app:related}

\begin{table}[ht]
\centering\fontsize{8}{10}\selectfont\setlength{\tabcolsep}{3pt}
\caption{Recruitment resources by language, outputs, candidate conditions, and label
 evidence. WorkRB coverage includes the accompanying toolkit \citep{workrb2026software}.}
\label{tab:related-benchmarks}
\begin{tabular}{@{}P{.20\textwidth}P{.15\textwidth}P{.29\textwidth}P{.31\textwidth}@{}}
\toprule
\rowcolor{EHRInk!5}Resource & Language / focus & Outputs / candidate conditions & Label evidence \\\midrule
ResumeBench\newline\mbox{\citep{ling2025resumebench}} & Five languages, including Chinese & Resume PDF to schema-defined JSON & Curated templates; GPT-4o-generated JSON aligned with PDF layouts \\
\addlinespace[3pt]
JobResQA\newline\mbox{\citep{carrino2026jobresqa}} & Five languages, including Chinese & Short answers and explanations over synthetic resume--JD pairs & Linguist-authored QA; human-assisted translation \\
\addlinespace[3pt]
PJB\newline\mbox{\citep{wang2026pjb}} & Chinese retrieval & 297 JD queries; fixed collection of 197,674 resumes & Doubao screening and Kimi relevance assessment; human checks on about 20\% of queries \\
\addlinespace[3pt]
WorkRB\newline\mbox{\citep{delange2026workrb}} & Multilingual work domain & Ranking and job-to-skill multilabel classification; ontology- and corpus-based candidate spaces & Task-specific datasets and occupational ontologies; extensible framework \\
\addlinespace[3pt]
\ehrbench & Chinese recruitment & 12 operations; generated text, typed spans, category paths, binary relations, retrieval, complete permutations & Source-aware label verification; assisted checks and expert judgments for constructed relations \\
\bottomrule\end{tabular}
\end{table}

\ehrbench makes twelve enterprise-informed Chinese recruitment operations available
for public LLM evaluation. Task-specific inputs, expert-reviewed references, and
output requirements support comparison of decisions, explanations, relations, and
candidate selection. Researchers can use these shared protocols to test whether model and
prompting improvements carry across tasks and sources. Supplied pools define the
candidate conditions for shortlist selection.

ResumeAtlas \citep{heakl2024resumeatlas} supplies resume classification data;
SkillSpan \citep{zhang2022skillspan} and Chinese-SkillSpan \citep{li2026chineseskillspan}
provide English skill spans and Chinese ESCO-aligned competency spans, respectively.
ConFit v2 \citep{yu2025confitv2} studies hard-negative retrieval training.

The WorkRB paper reports thirteen ranking tasks in seven groups. Its pinned toolkit
\citep{workrb2026software} also supports ESCO job-to-skill multilabel classification
and graded-relevance variants; our comparison includes these interfaces.

\begin{samepage}
Public/private comparisons also occur outside recruitment. UNED-ACCESS
\citep{sanchez2025unedaccess} compares previously unreleased entrance-exam questions
with a subject-matched subset of MMLU.
EHR-Bench uses its enterprise reference both to inform resource construction and to
contextualize the observed scores on a fixed shared-model set.\par
\end{samepage}

\FloatBarrier
\section{Reproducibility and Artifact Record}
\label{app:reproducibility}

The project repository is available at \url{https://github.com/EhrBench/EHR-Bench}.
The identifiers and commands below document the data, evaluation code, provenance
records, and per-sample scoring exports used in this study.

The corpus, prompts, and scorer are identified by commit \texttt{496964d}; saved
per-sample predictions and component statistics by \texttt{e8905f2}; and population
summary definitions by \texttt{b85bb3a}. The manifest records complete identifiers
and SHA-256 digests. All 144 scores for the twelve evaluated systems are recomputed
from the saved per-sample artifacts with the documented scoring formulas; every
model--task difference is at most 0.011 on the 0--100 scale. This check includes
binary labels, generation-component weights, strict-span aggregation, and candidate
metrics. It reuses saved rubric scores and does not independently rerun the judge.
Exported fields vary by task; generation exports provide component scores rather
than the complete raw response texts. Business-specific review procedures and
item-level expert records are not part of the public release.

From the manuscript directory, with the result archive available at
\texttt{/path/to/results}, the score check and table generation require:
\begin{quote}\footnotesize\begin{minipage}{\linewidth}
\texttt{python -m pip install -r scripts/requirements.txt}\\
\texttt{python scripts/validate\_v21.py --release-root \textbackslash}\\
\texttt{\ \ /path/to/results}\\
\texttt{python scripts/generate\_results.py}
\end{minipage}\end{quote}
The validator checks saved scorer and output artifacts; generated tables retain
six-family equal-weight Overall. The README gives commands for figures, paired
comparisons, candidate baselines, and the PJF probe. Auxiliary results record seeds,
features, entity groupings, software versions, and input hashes; additional PJF
dependencies are listed separately. Private-reference reproducibility uses aggregate
tables; raw records remain in the operational environment.

\end{document}